\documentclass[10pt,journal]{IEEEtran}

\usepackage{amsmath,amsfonts,amssymb,bm} 
\usepackage{pifont}          
\usepackage{tikz,graphicx,overpic}      
\usepackage{array,booktabs,multirow,makecell} 
\usepackage[table]{xcolor}              
\usepackage{arydshln}                   
\usepackage{ragged2e}        
\usepackage{algorithmic}    
\usepackage{enumerate}                  
\usepackage[compress]{cite}           
\usepackage[colorlinks,linkcolor=red,anchorcolor=blue,citecolor=green]{hyperref} 

\graphicspath{{./figs/}}
\DeclareGraphicsExtensions{.pdf,.jpeg,.png,.eps}

\newcommand{\eg}{\textit{e.g.}} 
\newcommand{\resubmit}[1]{{\textcolor{black}{#1}}}
\newcommand{\major}[1]{\textcolor{black}{#1}}
\begin{document}
\title{\resubmit{DRNet: All-in-One Image Restoration via Prior-Guided Dynamic Reparameterization}}

\author{
Ao~Li,~Xiaoning~Liu,~Sheng~Li,~Yapeng~Du,~Zhen~Long,~Lei~Luo,~Le Zhang,~and~Ce Zhu,~\IEEEmembership{Fellow,~IEEE}
\thanks{This work was supported by the Hainan Provincial Joint Project of Li'an International Education Innovation Pilot Zone under Grant 624LALH003, in part by the Key Program for International Cooperation of Ministry of Science and Technology of China under Grant 2024YFE0100700, and in part by the Key Project of the Natural Science Foundation of Sichuan Province under Grant 2025ZNSFSC0002. (Corresponding authors: Le Zhang; Ce Zhu.)}
\thanks{Ao Li, Xiaoning Liu, Sheng Li, Yapeng Du, Zhen Long, Le Zhang and Ce Zhu are with the School of Information and Communication Engineering, University of Electronic Science and Technology of China, Chengdu 611731, Sichuan, China.}
\thanks{Lei Luo is with the School of Communications and Information Engineering, Chongqing University of Posts and Telecommunication, Chongqing 400065, China.}
}

\markboth{Journal of \LaTeX\ Class Files,~Vol.~14, No.~8, August~2021}%
{Shell \MakeLowercase{\textit{et al.}}: A Sample Article Using IEEEtran.cls for IEEE Journals}


\maketitle

\begin{abstract}
\justifying
All-in-one image restoration aims to handle diverse degradations within a single model. However, existing methods often suffer from three key limitations: 1) per-input computational overhead from dynamic degradation estimation; 2) optimization challenges due to task heterogeneity; and 3) inefficient, frequency-agnostic encoder designs. To overcome these, we introduce the Dynamic Reparameterization Network (DRNet), a novel framework  operating on an initialization-stage reconfiguration paradigm that fundamentally eliminates per-input overhead. At its core, a Dynamic Reparameterization MLP (DRMLP) guided by a Task-Specific Modulator (TSM), which effectively mitigates task heterogeneity by orchestrating both specific restoration goals and a versatile general-purpose mode within a unified architecture. Furthermore, we incorporate a Continuous Wavelet Transform Encoder (CWTE) that explicitly leverages frequency characteristics via wavelet decomposition for a lightweight yet powerful design. \major{Extensive experiments demonstrate that DRNet achieves state-of-the-art performance across five restoration tasks with superior parameter efficiency.} Crucially, it showcases unique flexibility, excelling as both a highly competitive foundation model for blind restoration and a top-performing user-guided specialist. \major{The source code will be made available at: \url{https://github.com/AVC2-UESTC/DRNet-AiO.git}.}
\end{abstract}


\section{Introduction}
\label{sec:intro}
\begin{figure}[!t]
  \centering
   \includegraphics[width=\linewidth]{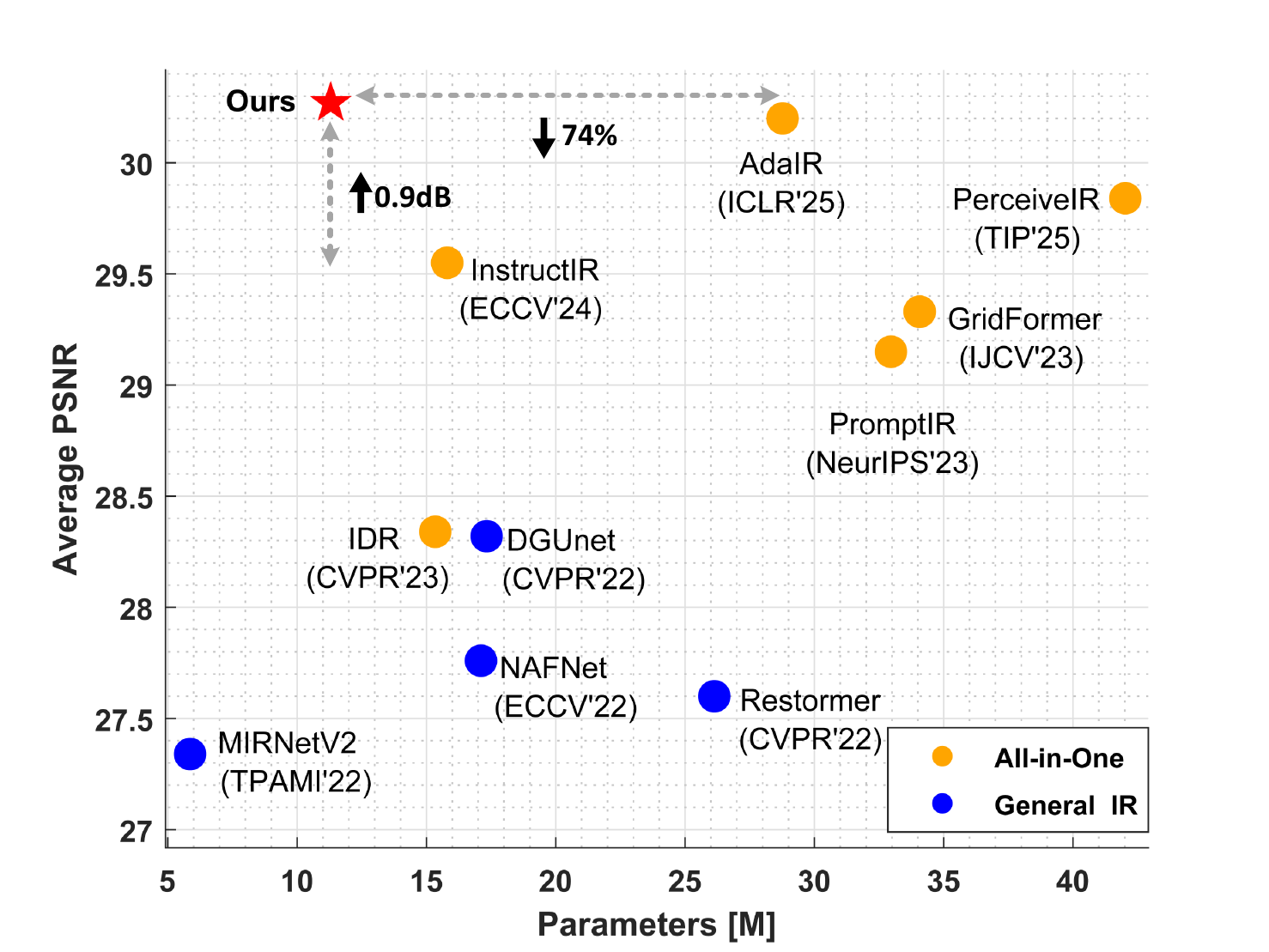}
    \caption{\resubmit{Comparison of our method against other approaches in terms of average PSNR and the number of parameters. The plot highlights our method's superior trade-off, achieving state-of-the-art performance with significantly fewer parameters. For instance, it reduces parameters by 74\% compared to AdaIR while improving PSNR by 0.9dB over InstructIR.}}
    \vspace{-5mm}
   \label{fig:psnr_param_cmp}
\end{figure}

\IEEEPARstart{I}{mage} restoration is a fundamental computer vision task aimed at reconstructing high-quality (HQ) images from their degraded counterparts. While traditional methods targeting specific degradations like denoising~\cite{liang2021swinir,wang2022uformer,zamir2022restormer}, deraining~\cite{yasarla2019uncertainty,liang2022drt}, and dehazing~\cite{liu2019griddehazenet,shin2021region} have achieved notable success, their limitation to singular degradation types curtails their utility in real-world scenarios where diverse distortions are prevalent. \major{This has spurred growing interest in all-in-one image restoration, a research direction that has been systematically analyzed in recent survey~\cite{jiang2024survey}, which aims to handle diverse degradations within a single framework and has achieved promising performance, as shown in Figure~\ref{fig:psnr_param_cmp}.}

\begin{figure}[!t]
  \centering
   \includegraphics[width=\linewidth]{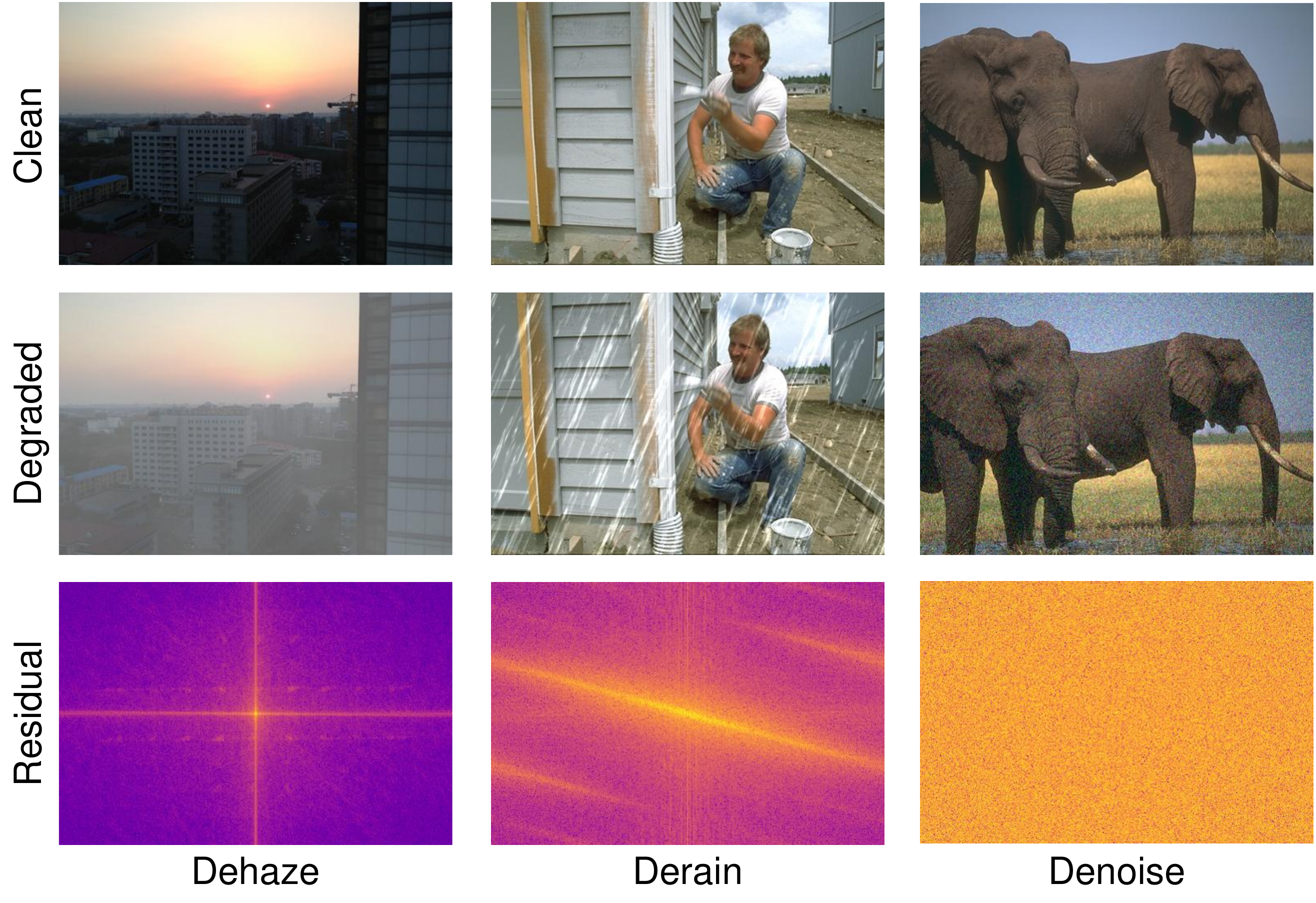}
   \caption{Different degradation impose different frequencies. The first row indicates the clean image, the second row represents the degraded image and the third row gives the spectrum of residual image which is obtained by using clean image subtract the degraded image. }
   \vspace{-6mm}
   \label{fig:preliminary}
\end{figure}

Despite these advancements, current all-in-one methods face three key limitations that hinder their practical applicability and efficiency: (1) A majority of existing methods rely on auxiliary modules or dedicated estimation networks to perform degradation estimation for each input image~\cite{conde2024instructir,li2023prompt,zhang2025perceive,liu2022tape,zhang2023ingredient}. While effective, this per-input estimation process often introduces substantial computational overhead during inference, making them less suitable for resource-constrained applications. (2) The inherent and considerable heterogeneity among different degradation types presents a major challenge for developing all-in-one models~\cite{zhang2025cvpr,yang2024all,kong2024towards}. This makes it difficult to learn task-specific knowledge effectively within a shared model. (3) Existing architectures often employ overly complex encoders, such as deep stacks of convolutional layers or generic transformer blocks, to learn robust features~\cite{valanarasu2022transweather,zhang2023ingredient,potlapalli2024promptir,wang2024gridformer,zhang2025perceive,cui2025adair,conde2024instructir}. While powerful, these encoders can be parameter-heavy and computationally intensive. More importantly, they do not fully exploit the intrinsic relationships between different degradations and how they distinctly alter image characteristics. As demonstrated in prior work such as AdaIR~\cite{cui2025adair}, frequency domain analysis provides valuable insights here. For example, Figure~\ref{fig:preliminary} shows that haze primarily affects low-frequency components, while noise impacts a broader range of frequencies, including high frequencies.

\resubmit{To overcome the aforementioned limitations, we introduce the Dynamic Reparameterization Network (DRNet), a novel and efficient framework for all-in-one image restoration that operates via initialization-stage network configuration. Unlike methods that rely on per-input degradation estimation, which introduces computational overhead, our approach configures the network based on a given prior before inference begins. This design allows our model to operate flexibly in two modes: (1) A general-purpose mode, which utilizes a generic, non-specific prior to handle unknown degradations, aligning with the traditional all-in-one paradigm. (2) A user-guided mode, where providing an explicit task specification leads to significant gains in restoration quality and eliminates the risk of misclassification errors.} 

\resubmit{To realize this dynamic configuration, we first address the challenges of per-input computational overhead and degradation heterogeneity. We propose the Dynamic Reparameterization MLP (DRMLP), a novel architecture inspired by structural reparameterization studies~\cite{ding2021repvgg,ding2022scaling,vasu2023mobileone}. The DRMLP employs multiple parallel pathways during training, which are guided by a lightweight Task-Specific Modulator (TSM). The TSM learns to generate adaptive weights based on the given prior—whether it is an explicit task instruction for user-guided mode, or a generic prior for general-purpose mode. During inference initialization, these weights dynamically fuse the parallel pathways into a single, efficient MLP tailored for the intended mode.}

Furthermore, to address the inefficiency and frequency-agnostic nature of conventional encoders, DRNet incorporates the Continuous Wavelet Transform Encoder (CWTE). Unlike traditional deep feature extractors, CWTE provides an explicit and interpretable multi-frequency decomposition via wavelet sub-band analysis~\cite{liu2024pasta,dong2023multi}. This explicit decomposition offers two significant advantages: (i) It facilitates a more targeted handling of degradation-specific frequency components, potentially leading to more effective restoration, and (ii) it significantly simplifies the encoder architecture. This leads to improved parameter efficiency without sacrificing representation power (Figure~\ref{fig:psnr_param_cmp}).

\resubmit{In summary, the main contributions of this work are: (1) We propose DRNet, a new framework that introduces an initialization-stage reconfiguration paradigm for all-in-one restoration. This fundamentally eliminates the per-input computational overhead of existing methods while offering unprecedented flexibility through both user-guided and general-purpose modes. (2) We introduce the Task-Specific Modulator (TSM), a unified and lightweight mechanism that resolves the core challenge of degradation heterogeneity. By generating adaptive weights from either specific or generic priors, it enables a single network to seamlessly transform into a specialized expert or a versatile generalist. (3) We design the Continuous Wavelet Transform Encoder (CWTE), a lightweight architecture that explicitly and interpretably leverages frequency characteristics. This design simultaneously enhances task-relevant feature representation and drastically simplifies the encoder structure, achieving superior performance with significantly fewer parameters.}

\section{Related Works}
\label{sec:relatedWorks}

\noindent\major{\textbf{General Image Restoration.} Modern image restoration has evolved from CNNs to Transformers and MLPs. SwinIR~\cite{liang2021swinir} and Restormer~\cite{zamir2022restormer} established the efficacy of attention mechanisms for spatial and channel-wise feature aggregation. Subsequent works like GRL~\cite{li2023efficient} and OKNet~\cite{cui2024omni} further improved multi-scale representation. Challenging the dominance of attention, MAXIM~\cite{tu2022maxim} demonstrated that pure MLP architectures could also model global interactions effectively. Recently, diffusion-based models~\cite{wu2024one,li2024light} have achieved photorealistic results but suffer from high computational costs. While these backbones provide powerful feature extraction, they lack the specific mechanisms required to handle the heterogeneity of all-in-one restoration tasks.}

\major{\noindent\textbf{All-in-One Image Restoration.} To handle diverse degradations, existing methods typically employ guiding mechanisms~\cite{li2022all,zhang2023ingredient,park2023all,conde2024instructir,potlapalli2024promptir,luo2024controlling,cheng2024continual,wen_allinone_tmm25}. Early approaches utilized degradation estimation via contrastive learning~\cite{li2022all} or physics-based models~\cite{zhang2023ingredient}. More recent strategies involve dynamic modulation, such as filter adaptation~\cite{park2023all} or prompt-based guidance~\cite{potlapalli2024promptir,conde2024instructir}. Recent advances also extend this unified paradigm to specific domains; for instance, Ada4DIR~\cite{lihe2025ada4dir} targets all-in-one remote sensing restoration, while UniUIR~\cite{zhang2025uniuir} considers underwater image restoration as an all-in-one learner. Notably, dynamic network designs have gained attention. For instance, Luo et al.~\cite{luo2023restoration} proposed a semi-dynamic network using per-input distortion embeddings. However, such per-input modulation introduces inference latency. In contrast, our DRNet adopts an {initialization-stage reconfiguration} strategy via the Task-Specific Modulator (TSM), ensuring zero overhead during inference execution while maintaining high adaptability.}

\begin{figure*}[!t]
  \centering
   \includegraphics[width=0.95\linewidth]{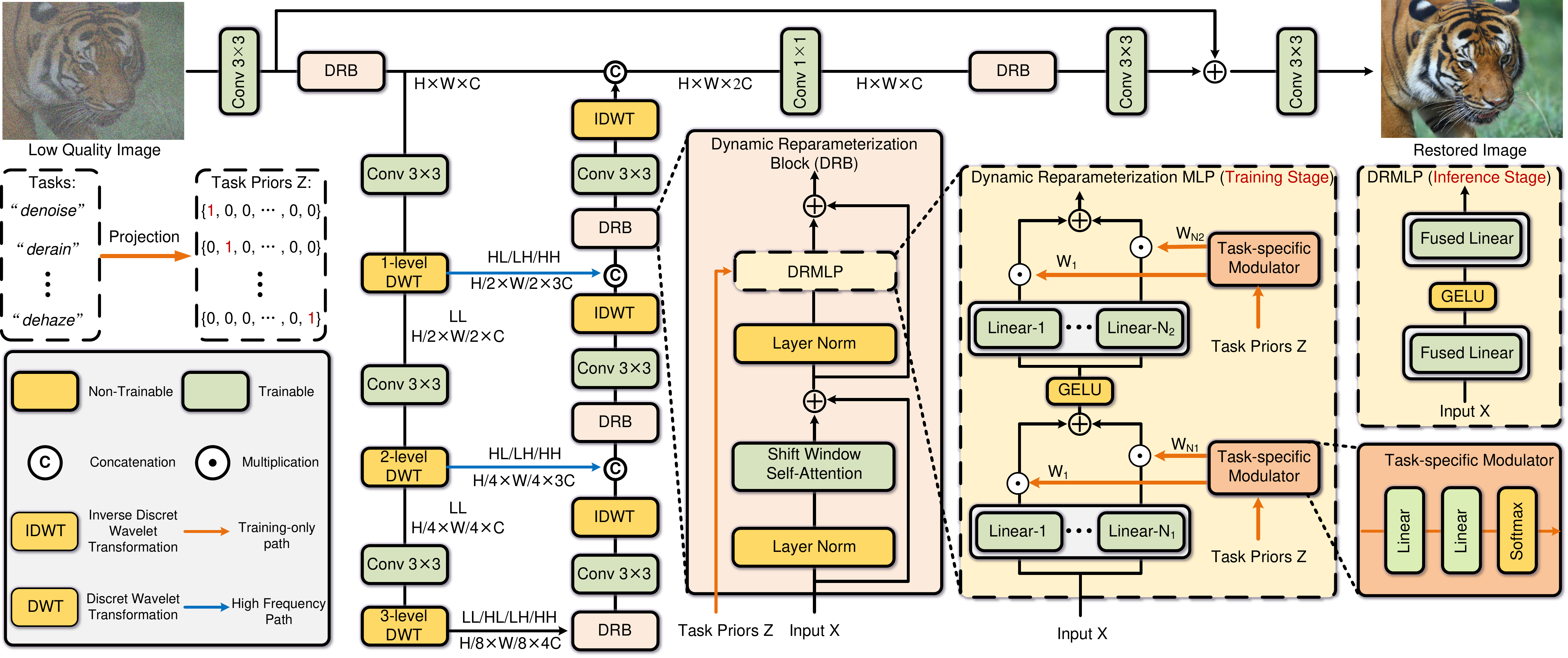}
   \caption{Overview of the Dynamic Reparameterization Network (DRNet). DRNet features a Continuous Wavelet Transform Encoder (CWTE) for efficient, frequency-aware feature extraction. The core innovation is the Dynamic Reparameterization MLP (DRMLP) which guided by a Task-Specific Modulator (TSM) undergoes a one-time dynamic reparameterization at inference initialization. These weights dynamically reparameterize the DRMLP, effectively creating a task-specific MLP structure for targeted restoration, eliminating the need for per-input dynamic adjustments.}
   \vspace{-5mm}
   \label{fig:framework}
\end{figure*}

\major{\noindent\textbf{Structural Reparameterization.} This technique decouples training-time multi-branch structures from inference-time single-branch architectures to boost efficiency~\cite{ding2021repvgg,ding2022scaling,vasu2023mobileone,zhang2021edge,yan2025mobileie}. For example, RepVGG~\cite{ding2021repvgg} merges parallel convolutional branches into a single kernel for inference. ECBSR~\cite{zhang2021edge} introduces an edge-oriented convolution layer into a series of parallel convolution layers, then fuses them during the inference stage, achieving significant performance. MobileIE~\cite{yan2025mobileie} expands more convolution layers in the parallel structure and adds the Batch Normalization layer in each branch, further boosting performance. However, these methods typically perform {static} fusion for single-task models. They are ill-suited for all-in-one restoration where the optimal weights vary across tasks. Our work pioneers {dynamic, prior-guided reparameterization}. Unlike static fusion, our DRMLP reconfigures its pathways based on the task prior, creating specialized experts from a universal architecture to resolve task interference.}

\major{\noindent\textbf{Frequency-aware Image Restoration.} Frequency analysis enables effective representation learning by separating spectral components. In single-task restoration, methods have utilized frequency masking~\cite{helou2020stochastic}, recursive residual modules~\cite{qiu2019embedded}, DCT-based partitioning~\cite{xie2021learning}, or implicit modulation modules~\cite{li2025exploring} to enhance high-frequency recovery. Dynamic filtering in the frequency domain has also been explored for super-resolution~\cite{magid2021dynamic,zhong2018joint}. Beyond standard image restoration, frequency learning has also proven effective in complex video tasks. For example, ProFiT~\cite{chen2025profit} introduces a prompt-guided frequency-aware filtering framework for hyperspectral video tracking, further confirming the robustness of frequency domain representations in handling temporal and spectral data.}

\major{In the all-in-one setting, recent works like HFRSN~\cite{liu2021multi}, FSDGN~\cite{yu2022frequency}, and AdaIR~\cite{cui2025adair} leverage frequency mining to guide the network. However, most methods rely on complex attention mechanisms or implicit learning for frequency selection. In contrast, our DRNet introduces the Continuous Wavelet Transform Encoder (CWTE) to {explicitly} leverage the physical properties of wavelets. By using parameter-free wavelet downsampling, CWTE efficiently disentangles degradation patterns without the computational burden of heavy learnable modules.}

\section{Methodology}
We propose the Dynamic Reparameterization Network (DRNet), with its overall architecture illustrated in Figure~\ref{fig:framework}. Given an input low-quality (LQ) image $\boldsymbol{X}_{lq}\in\mathbb{R}^{3\times H\times W}$, the network first applies a convolutional layer to extract shallow features $\boldsymbol{F}_{s}\in\mathbb{R}^{C\times H\times W}$, where $H\times W$ represent the spacial dimension and $C$ indicates the number of features. These shallow features are initially processed by a Dynamic Reparameterization Block (DRB) before being fed into a four-level hourglass-shaped structure. Within this structure, we incorporate the continuous wavelet transform to decompose shallow features. Each level of the decoder consists of multiple DRBs, which progressively refine the features. The outputs from the decoder are concatenated with the output of the first DRB to obtain deep features $\boldsymbol{F}_d\in\mathbb{R}^{2C\times H\times W}$, followed by a $1\times1$ convolutional layer to reduce the channel dimensions. Subsequently, an additional DRB refines the fuse features, which is then processed by a $3\times3$ convolutional layer. Finally, we establish a long skip connection between the shallow and refined features, applying another $3\times3$ convolutional layer to generate the restored image $\boldsymbol{X}_{r}\in\mathbb{R}^{3\times H\times W}$.

\subsection{\resubmit{Prior-Guided Network Reconfiguration}}
\label{subsec:classificationAnalysis}
\major{Unlike existing frameworks that introduce computational overhead by processing degradation priors for each input instance~\cite{conde2024instructir,potlapalli2024promptir}, we propose an initialization-stage reconfiguration paradigm. This approach configures the network parameters based on the task prior \textit{before} inference execution, ensuring zero per-image overhead.}

\resubmit{The process begins by mapping a conceptual restoration task, $\mathcal{T}$, into a task embedding vector $\boldsymbol{Z}$ using a simple, predefined projector $\phi(\cdot)$:
\begin{equation}
  \boldsymbol{Z} = \phi(\mathcal{T}).
\label{eq:addeq1}
\end{equation}
\major{Formally, let the set of supported tasks be indexed by $k \in \{1, \dots, K\}$. The projector $\phi(\cdot)$ maps the $k$-th task to a fixed standard basis vector (one-hot vector) $\boldsymbol{Z} \in \{0,1\}^K$. While these input vectors are mutually orthogonal, the TSM employs \textit{shared learnable parameters} to process them. Through joint training, the TSM learns a mapping function that transforms these distinct one-hot inputs into specific combinations of weighting coefficients ($w$) for the DRMLP. This mechanism allows the model to discover cross-task relationships: for semantically similar tasks, the TSM learns to generate similar weight distributions, thereby activating the underlying feature extractors in a coordinated manner (as analyzed in Section~\ref{ab:analysis_TSM}).} This flexible formulation supports DRNet's two operational modes by interpreting different types of tasks for $\mathcal{T}$: (1) User-Guided Mode: In this mode, $\mathcal{T}$ represents an explicit, user-provided instruction, such as ``denoise" or ``derain". The projector $\phi(\cdot)$ maps this instruction to a corresponding one-hot vector. This allows for highly specialized network configurations that maximize restoration quality and eliminate any risk of misclassification. (2) General-Purpose (Blind) Mode: When no specific user guidance is available, $\mathcal{T}$ is set to a generic ``blind" instruction. The projector $\phi(\cdot)$ treats this as another distinct task and maps it to its own pre-defined one-hot vector. This vector configures the network into a versatile state, capable of handling unknown degradations.}

\resubmit{In our primary experiments, we focus on the User-Guided Mode to demonstrate the upper bounds of DRNet's performance and efficiency. To validate the model's versatility, we also provide results for the General-Purpose Mode, which are labeled ``Ours w/o prompt".}

\subsection{Dynamic Reparameterization Block}
\label{sec:dmb}
The Dynamic Reparameterization Block (DRB) serves as the fundamental feature processing unit in DRNet, adapting the architecture of the Swin Transformer block~\cite{Liu2021iccv}. As illustrated in Figure~\ref{fig:framework}, each DRB comprises two main components: a Shift-Window Self-Attention (SW-SA) and our novel Dynamic Reparameterization MLP (DRMLP).
\begin{figure}[!t]
  \centering
   \includegraphics[width=\linewidth]{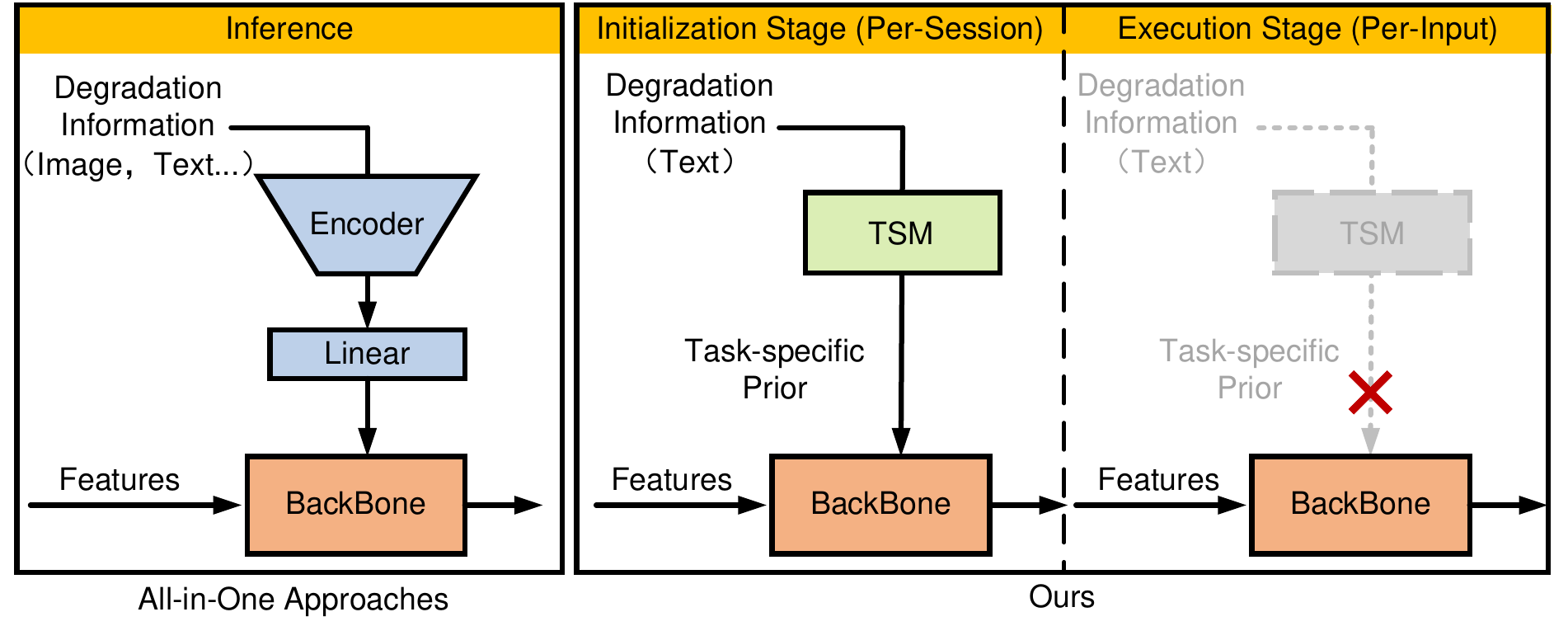}
\caption{\major{Comparison of prior integration architectures. Existing all-in-one models (left) typically rely on dedicated modules that actively process degradation information for {each input image}, introducing constant computational overhead during inference. In contrast, our approach (right) introduces an {Initialization-Stage Reconfiguration} paradigm. The network is configured {once per task session} (not per input) by fusing the dynamic weights into a static backbone. This mechanism completely eliminates prior-related computation during the actual image processing (Execution Stage), ensuring {zero per-input overhead}.}}
\vspace{-5mm}
\label{fig:control_cmp}
\end{figure}
\subsubsection{Shift-Window Self-Attention}
\major{To capture global spatial information while maintaining computational efficiency, we adopt the standard Shift-Window Self-Attention (SW-SA) mechanism following Swin Transformer~\cite{Liu2021iccv}. 
Given the input feature $\boldsymbol{F}_i$, the SW-SA module partitions it into non-overlapping local windows and performs self-attention within each window. To enable cross-window connections, the partitioning configuration is shifted cyclically in consecutive blocks. We incorporate relative position embeddings and Layer Normalization (LN) consistent with the standard implementation. Since this module is not the focus of our dynamic reparameterization, we refer readers to~\cite{Liu2021iccv} for detailed mathematical formulations.}
\subsubsection{Dynamic Reparameterization MLP} 
The Dynamic Reparameterization MLP (DRMLP) comprises two main components: a Multi-Branch Parallel Architecture and a Task-Specific Modulator (TSM).

\textbf{Multi-Branch Parallel Architecture.} The DRMLP employs a two-stage structure with banks of parallel linear layers. The first stage consists of $N_1$ linear layers, $Bank_1 = \{linear_1^1, \dots, linear_1^{N_1}\}$, and the second stage consists of $N_2$ linear layers, $Bank_2 = \{linear_2^1, \dots, linear_2^{N_2}\}$. During training, the transformation of $\boldsymbol{\hat{F}}_i^{attn}$ (the input feature after normalization) is formulated as:
\begin{equation}
  \boldsymbol{F}_{i} = \sum^{N_2}_{k=1}{w^{k}_{2} \cdot linear^{k}_{2}\left(\sigma\left(\sum^{N_1}_{j=1}{w^{j}_{1} \cdot linear^{j}_{1}(\boldsymbol{\hat{F}}_i^{attn})}\right)\right)},
\label{eq:drmlp_train}
\end{equation}
where $\sigma(\cdot)$ is an activation function (\eg, GELU). The weight coefficients are dynamically computed by the TSM, as detailed below.

\resubmit{\textbf{Task-Specific Modulator.} The TSM is the core component that enables the dynamic adaptation of the DRMLP. Its function is to translate a high-level task prior embedding into low-level architectural weights $\boldsymbol{W}_1 \in \mathbb{R}^{N_1}$ and $\boldsymbol{W}_2 \in \mathbb{R}^{N_2}$. We define a ``task" broadly to include not only specific restoration goals (\eg, ``denoise") but also a versatile ``general-purpose" task for blind restoration.} \major{Each task is strictly represented by a unique fixed one-hot vector $\boldsymbol{Z}$ (as defined in Eq.~\ref{eq:addeq1}).} Given the embedding $\boldsymbol{Z}$ for the intended task, the TSM processes it through shared linear layers to compute the weight coefficient vectors $\boldsymbol{W}_1$ and $\boldsymbol{W}_2$ as:
\begin{align}
  \boldsymbol{W}_{1} &= \text{Softmax}(linear_{1b}(linear_{1a}(\boldsymbol{Z}))), \label{eq:tsm_w1} \\
  \boldsymbol{W}_{2} &= \text{Softmax}(linear_{2b}(linear_{2a}(\boldsymbol{Z}))), \label{eq:tsm_w2}
\end{align}
where ${linear}_{xa}$ and ${linear}_{xb}$ represent the linear layers within the TSM. \major{These generated weights $w_1^j \in \boldsymbol{W}_1$ and $w_2^k \in \boldsymbol{W}_2$ are then used to modulate the parallel branches in Eq.~\ref{eq:drmlp_train}.}

\textbf{Inference-Time Reparameterization.} A key feature of DRMLP is its ability to be reparameterized into a more efficient structure during inference initialization. The parallel linear layers within each bank can be aggregated. Specifically, for $Bank_1$, its effective operation on $\boldsymbol{\hat{F}}_i^{attn}$ becomes:
\begin{equation}
  \boldsymbol{F}^{1}_{i} = \sum^{N_1}_{j=1}{w^{j}_{1} \cdot linear^{j}_{1}(\boldsymbol{\hat{F}}_i^{attn}}).
\label{eq:bank1_sum}
\end{equation}
Leveraging the linearity of this weighted sum, $Bank_1$ can be fused into a single equivalent linear layer, $linear_1^{new}$:
\begin{equation}
\begin{split}
  \boldsymbol{F}^{1}_{i} &= \sum^{N_1}_{j=1}{w^j_1 \cdot (\boldsymbol{W}^{j,linear}_{1} \cdot \boldsymbol{\hat{F}}_i^{attn} + \boldsymbol{b}^{j,linear}_{1})}\\
&= \left(\sum^{N_1}_{j=1}{w^j_{1} \cdot \boldsymbol{W}^{j,linear}_{1}}\right) \cdot \boldsymbol{\hat{F}}_i^{attn} + \sum^{N_1}_{j=1}{(w^j_{1} \cdot \boldsymbol{b}^{j,linear}_{1})} \\
&= \boldsymbol{W}^{new}_{1} \cdot \boldsymbol{\hat{F}}_i^{attn} + \boldsymbol{b}^{new}_{1} = linear_1^{new}(\boldsymbol{\hat{F}}_i^{attn}).
\end{split}
\label{eq:bank1_reparam}
\end{equation}
Similarly, $Bank_2$ can be reparameterized into an equivalent $linear_2^{new}$ using weights $w_2^k$. \major{To address potential ambiguities regarding the source of weights $w$ during inference, we explicitly distinguish between two operational stages. {(1) Initialization Stage (Per-Session):} This stage occurs once when the restoration task is defined (e.g., when the user selects ``denoise" mode). The TSM computes the scalar weights $w$ based on the prior $\boldsymbol{Z}$, and the parallel branches are mathematically fused into $linear^{new}_{1,2}$ using Eq.~(9). {(2) Execution Stage (Per-Image):} During the actual image processing, the network operates as a static model using the pre-fused $linear^{new}_{1,2}$. The TSM and the prior $\boldsymbol{Z}$ are no longer involved. Thus, although Eq.~(9) mathematically involves $w$ (derived from the prior), these values are absorbed into the static weights $\boldsymbol{W}^{new}_{1,2}$ before any image is processed, ensuring zero per-input computational overhead as claimed in Figure~{\ref{fig:control_cmp}.}}

\major{\textbf{Attention Reparameterization.} While dynamic reparameterization could theoretically be applied to the SW-SA module, we deliberately limit it to the MLP block. The SW-SA module already dominates computational cost, and adding parallel branches for Q, K, V projections during training would prohibitively increase GPU memory usage. Focusing reparameterization on the MLP offers the best trade-off between adaptability and training feasibility.}

\major{\textbf{Difference from Adapters and Dynamic Convolution.} Conventional adapters~\cite{houlsby2019parameter} insert additional layers into the network, which increases FLOPs and latency for every inference pass. Similarly, dynamic convolutions~\cite{chen2020dynamic} typically generate weights conditioned on each input image, incurring per-input computational overhead. In contrast, DRNet adopts an {initialization-stage reconfiguration} strategy. Since our modulation is conditioned on the task prior rather than the input image, the dynamic weights are computed and fused into a static structure {only once} when the task mode is selected. Consequently, during the actual processing of images, DRNet operates with {zero additional overhead} compared to a static baseline, ensuring maximum efficiency.}
\subsection{Continuous Wavelet Transform Encoder}
\label{subsec:spectrumAnalysis}
\major{To explicitly leverage frequency-domain characteristics for efficient feature disentanglement, we introduce the Continuous Wavelet Transform Encoder (CWTE). Unlike conventional encoders that rely on deep stacks of learnable downsampling layers, CWTE utilizes the discrete Haar wavelet decomposition to hierarchically separate features with minimal computational cost.}

\major{Formally, at each encoder level, the input feature map $\boldsymbol{X}_{in} \in \mathbb{R}^{C\times H\times W}$ first undergoes local refinement via a $3\times3$ convolution. Subsequently, Wavelet Decomposition (WD) explicitly separates the features into four frequency sub-bands:
\begin{equation}
\{\boldsymbol{LL}, \boldsymbol{LH}, \boldsymbol{HL}, \boldsymbol{HH}\} = \text{WD}(f_{3\times3}(\boldsymbol{X}_{in})).
\label{eq:dwt_decomposition_explicit}
\end{equation}
We implement a frequency-aware routing strategy based on these sub-bands. The high-frequency components ($\boldsymbol{LH}, \boldsymbol{HL}, \boldsymbol{HH}$), which typically capture fine details and degradation artifacts (\eg, noise or rain streaks), are directly routed to the corresponding decoder stage via skip connections. This preserves high-frequency fidelity without the risk of distortion from deeper network layers. In contrast, the low-frequency sub-band ($\boldsymbol{LL}$), which carries the core structural content, is passed to the next encoder level for recursive decomposition. This design constructs a lightweight hierarchical encoder that effectively disentangles degradation patterns from image structure.}
\begin{table*}[!t]
  \centering
  \footnotesize 
 \caption{\resubmit{Performance comparison of general image restoration and all-in-one methods on five tasks. ``Params" represents the total number of network parameters. The best and second-best results for each setting are {\textbf{highlighted}} and {\underline{underlined}}, respectively. This table is based on Perceive-IR~\cite{zhang2025perceive} (TIP ’25).}}
  \setlength{\tabcolsep}{1mm}
  \begin{tabular}{lccccccccccccc}
    \toprule
    & & \multicolumn{2}{c}{\textbf{Derain}} & \multicolumn{2}{c}{\textbf{Dehaze}} & \multicolumn{2}{c}{\textbf{Denoise}} & \multicolumn{2}{c}{\textbf{Deblur}} & \multicolumn{2}{c}{\textbf{Low-light Enh.}} &  &  \\
    
     \multicolumn{1}{c}{\bf Methods} & \textbf{Params} & \multicolumn{2}{c}{Rain100L~\cite{yang2017deep}} & \multicolumn{2}{c}{SOTS~\cite{li2018benchmarking}} & \multicolumn{2}{c}{BSD68$_{\sigma=25}$~\cite{martin2001database}} & \multicolumn{2}{c}{GoPro~\cite{nah2017deep}} & \multicolumn{2}{c}{LOL~\cite{wei2018deep}} & \multicolumn{2}{c}{\textbf{Average}} \\
     
    \cmidrule(lr){3-4}\cmidrule(lr){5-6}\cmidrule(lr){7-8}\cmidrule(lr){9-10}\cmidrule(lr){11-12}\cmidrule(lr){13-14}
    & (M) & PSNR$\uparrow$ & SSIM$\uparrow$ & PSNR$\uparrow$ & SSIM$\uparrow$ & PSNR$\uparrow$ & SSIM$\uparrow$ & PSNR$\uparrow$ & SSIM$\uparrow$ & PSNR$\uparrow$ & SSIM$\uparrow$ & PSNR$\uparrow$ & SSIM$\uparrow$ \\
    \midrule
    \midrule
    \multicolumn{14}{c}{\textit{\textbf{\textcolor{blue}{General Image Restorers}}}} \\
    (TPAMI'22) MIRNetV2~\cite{zamir2020learning}  & 5.86 & 33.89 & 0.954 & 24.03 & 0.927 & 30.97 & 0.881 & 26.30 & 0.799 & 21.52 & 0.815 & 27.34 & 0.875 \\
    (CVPR'22) DGUNet~\cite{mou2022deep} & 17.33 & 36.62 & 0.971 & 24.78 & 0.940 & 31.10 & 0.883 & 27.25 & 0.837 & 21.87 & 0.823 & 28.32 & 0.891 \\
    (CVPR'22) Restormer~\cite{zamir2022restormer} & 26.13 & 34.81 & 0.962 & 24.09 & 0.927 & 31.49 & 0.884 & 27.22 & 0.829 & 20.41 & 0.806 & 27.60 & 0.881 \\
    (ECCV'22) NAFNet~\cite{chen2022simple} & 17.11 & 35.56 & 0.967 & 25.23 & 0.939 & 31.02 & 0.883 & 26.53 & 0.808 & 20.49 & 0.809 & 27.76 & 0.881 \\
    (TPAMI'23) FSNet~\cite{cui2023image} & 13.28 & 36.07 & 0.968 & 25.53 & 0.943 & 31.33 & 0.883 & 28.32 & 0.869 & 22.29 & 0.829 & 28.71 & 0.898 \\
    (ECCV'24) MambaIR~\cite{guo2024mambair} & 26.78 & 36.55 & 0.971 & 25.81 & 0.944 & 31.41 & 0.884 & 28.61 & 0.875 & 22.49 & 0.832 & 28.97 & 0.901 \\
    \midrule
    \midrule
    \multicolumn{14}{c}{\textit{\textbf{\textcolor{red}{All-in-One Image Restorers}}}} \\
    (TPAMI'19) DL~\cite{fan2019general} & 2.09 & 21.96 & 0.762 & 20.54 & 0.826 & 23.09 & 0.745 & 19.86 & 0.672 & 19.83 & 0.712 & 21.05 & 0.743 \\
    (CVPR'22) T.weather~\cite{valanarasu2022transweather} & 37.93 & 29.43 & 0.905 & 21.32 & 0.885 & 29.00 & 0.841 & 25.12 & 0.757 & 21.21 & 0.792 & 25.22 & 0.836 \\
    (ECCV'22) TAPE~\cite{liu2022tape} & 1.07 & 29.67 & 0.904 & 22.16 & 0.861 & 30.18 & 0.855 & 24.47 & 0.763 & 18.97 & 0.621 & 25.09 & 0.801 \\
    (CVPR'22) AirNet~\cite{li2022all} & 8.93 & 32.98 & 0.951 & 21.04 & 0.884 & 30.91 & 0.882 & 24.35 & 0.781 & 18.18 & 0.735 & 25.49 & 0.846 \\
    (CVPR'23) IDR~\cite{zhang2023ingredient} & 15.34 & 35.63 & 0.965 & 25.24 & 0.943 & {\textbf{31.60}} & {0.887} & {27.87} & {0.846} & 21.34 & 0.826 & 28.34 & 0.893 \\
    (NeurIPS'23) PromptIR~\cite{potlapalli2024promptir} & 32.96 & 36.37 & 0.970 & 26.54 & 0.949 & 31.47 & 0.886 & 28.71 & 0.881 & 22.68 & 0.832 & 29.15 & 0.904 \\
    (IJCV'24) Gridformer~\cite{wang2024gridformer} & 34.07 & 36.61 & 0.971 & 26.79 & 0.951 & 31.45 & 0.885 & 29.22 & {\underline{0.884}} & 22.59 & 0.831 & 29.33 & 0.904 \\
    
     (ECCV'24) InstructIR-5D~\cite{conde2024instructir} & 15.8 & 36.84 & 0.973 & 27.10 & 0.956 & 31.40 & {0.887} & {\underline{29.40}} & {\textbf{0.886}} & {\textbf{23.00}} & 0.836 & 29.55 & 0.907 \\

    (TIP'25) Perceive-IR~\cite{zhang2025perceive} & 42.02 &{37.25} & {0.977} & \underline{28.19} & {0.964} & 31.44 & {0.887} & {\textbf{29.46}} & {\textbf{0.886}} & {\underline{22.88}} & 0.833 & 29.84 & 0.909 \\
    
    (ICLR'25) AdaIR~\cite{cui2025adair} & 28.77 & {\underline{38.02}} & {\underline{0.981}} & {{30.53}} & {\underline{0.978}} & 31.35 & {\underline{0.889}} & 28.12 & 0.858 & {\textbf{23.00}} & {\underline{0.845}} & {\underline{30.20}} & {\underline{0.910}} \\

    \rowcolor{gray!20} \multicolumn{1}{c}{Ours w/o prompt} & 7.39 & 36.57 & 0.977 & \underline{30.66} & \underline{0.978} & 31.37 & \underline{0.890} & 27.52 & 0.831 & 21.99 & 0.841 & 29.62 & 0.903 \\
    \rowcolor{gray!20} \multicolumn{1}{c}{Ours} & 7.39 & {\textbf{38.13}} & {\textbf{0.982}} & {\textbf{31.28}} & {\textbf{0.980}}  & {\underline{31.54}} & {\textbf{0.894}} & 29.01 & 0.870 & {22.30} & {\textbf{0.846}} & {\textbf{30.45}} & {\textbf{0.914}} \\
    
    \bottomrule
  \end{tabular}
  \vspace{-4mm}
  \label{tab:5tasks}
\end{table*}
\begin{table*}[!t]
  \centering
  \footnotesize 
 \caption{\major{Performance comparison of general image restoration and all-in-one methods on three tasks. ``Params" represents the total number of network parameters. The best and second-best results for each setting are \textbf{highlighted} and \underline{underlined}, respectively. This table is based on Perceive-IR~\cite{zhang2025perceive} (TIP ’25).}}
  \setlength{\tabcolsep}{1mm}
  \begin{tabular}{lccccccccccccc}
    \toprule
    & & \multicolumn{2}{c}{\textbf{Derain}} & \multicolumn{2}{c}{\textbf{Dehaze}} & \multicolumn{6}{c}{\textbf{Denoise (BSD68~\cite{martin2001database})}} &  &  \\
    \cmidrule(lr){7-12}
     \multicolumn{1}{c}{\bf Methods} & \textbf{Params} & \multicolumn{2}{c}{Rain100L~\cite{yang2017deep}} & \multicolumn{2}{c}{SOTS~\cite{li2018benchmarking}} & \multicolumn{2}{c}{${\sigma=15}$} & \multicolumn{2}{c}{${\sigma=25}$} & \multicolumn{2}{c}{${\sigma=50}$} & \multicolumn{2}{c}{\textbf{Average}} \\
     
    \cmidrule(lr){3-4}\cmidrule(lr){5-6}\cmidrule(lr){7-8}\cmidrule(lr){9-10}\cmidrule(lr){11-12}\cmidrule(lr){13-14}
    & (M) & PSNR$\uparrow$ & SSIM$\uparrow$ & PSNR$\uparrow$ & SSIM$\uparrow$ & PSNR$\uparrow$ & SSIM$\uparrow$ & PSNR$\uparrow$ & SSIM$\uparrow$ & PSNR$\uparrow$ & SSIM$\uparrow$ & PSNR$\uparrow$ & SSIM$\uparrow$ \\
    \midrule
    \midrule
    \multicolumn{14}{c}{\textit{\textbf{\textcolor{blue}{General Image Restorers}}}} \\
    (CVPR'21) MPRNet~\cite{zamir2021multi} & 17.33 & 33.86 & 0.958 & 28.00 & 0.958 & 33.27 & 0.920 & 30.76 & 0.871 & 27.29 & 0.761 & 30.63 & 0.894 \\
    (CVPR'22) Restormer~\cite{zamir2022restormer} & 26.13 & 33.78 & 0.958 & 27.78 & 0.958 & 33.72 & 0.930 & 30.67 & 0.865 & 27.63 & 0.792 & 30.75 & 0.901 \\
    (ECCV'22) NAFNet~\cite{chen2022simple} & 17.11 & 33.64 & 0.956 & 24.11 & 0.928 & 33.03 & 0.918 & 30.47 & 0.865 & 27.12 & 0.754 & 29.67 & 0.844 \\
    (TPAMI'23) FSNet~\cite{cui2023image} & 13.28 & 35.61 & 0.969 & 29.14 & 0.968 & 33.81 & 0.930 & 30.84 & 0.872 & 27.69 & 0.792 & 31.42 & 0.906 \\
    (ECCV'24) MambaIR~\cite{guo2024mambair} & 26.78 & 35.42 & 0.969  & 29.57 & 0.970 & 33.88 & 0.931 & 30.95 & 0.874 & 27.74 & 0.793 & 31.51 & 0.907 \\
    \midrule
    \midrule
    \multicolumn{14}{c}{\textit{\textbf{\textcolor{red}{All-in-One Image Restorers}}}} \\
    (CVPR'22) AirNet~\cite{li2022all} & 8.93 & 34.90 & 0.967 & 27.94 & 0.962 & 33.92 & 0.932 & 31.26 & 0.888 & 28.00 & 0.797 & 31.20 & 0.910 \\
    (CVPR'23) IDR~\cite{zhang2023ingredient} & 15.34 & 36.03 & 0.971 & 29.87 & 0.970 & 33.89 & 0.931 & 31.32 & 0.884 & 28.04 & 0.798 & 31.83 & 0.911\\
    (NeurIPS'23) PromptIR~\cite{potlapalli2024promptir} & 32.96 & 36.37 & 0.972 & 30.58 & 0.974 & 33.98 & 0.933 & 31.31 & 0.888 & 28.06 & 0.799 & 32.06 & 0.913  \\
    (IJCV'24) Gridformer~\cite{wang2024gridformer} & 34.07 & 37.15 & 0.972 & 30.37 & 0.970 & 33.93 & 0.931 & 31.37 & 0.887 & 28.11 & 0.801 & 32.19 & 0.912\\
    
     (ECCV'24) InstructIR-3D~\cite{conde2024instructir} & 15.8 & 37.98 & 0.978 & 30.22 & 0.959 & \underline{34.15} & 0.933 & 31.52 & {0.890} & \underline{28.30} & \underline{0.804} & 32.43 & 0.913 \\

    (TIP'25) Perceive-IR~\cite{zhang2025perceive} & 42.02 & \underline{38.29} & \underline{0.980} & {30.87} & {0.975} & 34.13 & {0.934} & {\underline{31.53}} & {{0.890}} & {\textbf{28.31}} & \underline{0.804} & \underline{32.63} & \underline{0.917} \\
    
    (ICLR'25) AdaIR~\cite{cui2025adair} & 28.77 & {\textbf{38.64}} & \textbf{0.983} & {{31.06}} & {\textbf{0.980}} & 34.12 & {\underline{0.935}} & 31.45 & \underline{0.892} & {{28.19}} & {{0.802}} & {\textbf{32.69}} & {\textbf{0.920}} \\

    \rowcolor{gray!20} \multicolumn{1}{c}{Ours w/o prompt} & 7.39 & 37.03 & 0.979 & \underline{31.07} & \underline{0.979} & 34.11 & \underline{0.935} & 31.44 & \underline{0.892} & 28.16 & 0.802 & 32.36 & \underline{0.917} \\
    \rowcolor{gray!20} \multicolumn{1}{c}{Ours} & 7.39 & 38.28 & {\textbf{0.983}} & {\textbf{31.15}} & {\underline{0.979}}  & {\textbf{34.20}} & {\textbf{0.937}} & \textbf{31.55} & \textbf{0.894} & {28.27} & {\textbf{0.807}} & {\textbf{32.69}} & {\textbf{0.920}} \\
    
    \bottomrule
  \end{tabular}
  \vspace{-4mm}
  \label{tab:3tasks}
\end{table*}

\section{Experiments}
\label{sec:exp} 
\subsection{Datasets}
For a fair comparison, we follow the experimental settings of prior works~\cite{potlapalli2024promptir,li2022all,cui2025adair}. For image denoising, we use a combination of BSD400~\cite{arbelaez2010contour} and WED~\cite{ma2016waterloo} for training. Noisy images are generated by adding white Gaussian noise to the clean images at three noise levels: $\sigma = 15, 25, 50$. For evaluation, we use BSD68~\cite{martin2001database}, Urban100~\cite{huang2015single}, and Kodak24~\cite{franzen1999kodak}. For image deraining, we utilize the Rain100L dataset~\cite{yang2017deep}. For image dehazing, we use the Outdoor Training Set (OTS)~\cite{li2018benchmarking}. For image deblurring, we train and evaluate on the GoPro dataset~\cite{nah2017deep}. Finally, for low-light image enhancement, we adopt the LOL dataset~\cite{wei2018deep}.

\subsection{Implementation Details}
We randomly crop images into 128 $\times$ 128 patches and train for a total of 150 epochs. Data augmentation, such as random horizontal flipping and 90$^\circ$ rotations, is applied to enhance generalization. The Adam optimizer is used to minimize the $\ell_1$ loss. The batch size is set to 32, and the initial learning rate is $2 \times 10^{-4}$. We also use cosine annealing learning rate decay. Our model employs a 4-level hourglass-shaped structure. At level 1, we apply the DRB with 4 blocks, 1 attention head, and 48 channels. At levels 2 to 4, DRB blocks are used to extract deep features, with the number of blocks set to [6, 6, 8], attention heads to [2, 4, 8], and channels to 192. The refinement stage consists of 4 DRB blocks. For all DRB blocks, the channel expansion factor is 2, and the window size is 8. Additionally, the number of linear layers in both $Bank_1$ and $Bank_2$ is set to 4.
\begin{figure*}[!t]
  \centering
   \includegraphics[width=0.95\linewidth]{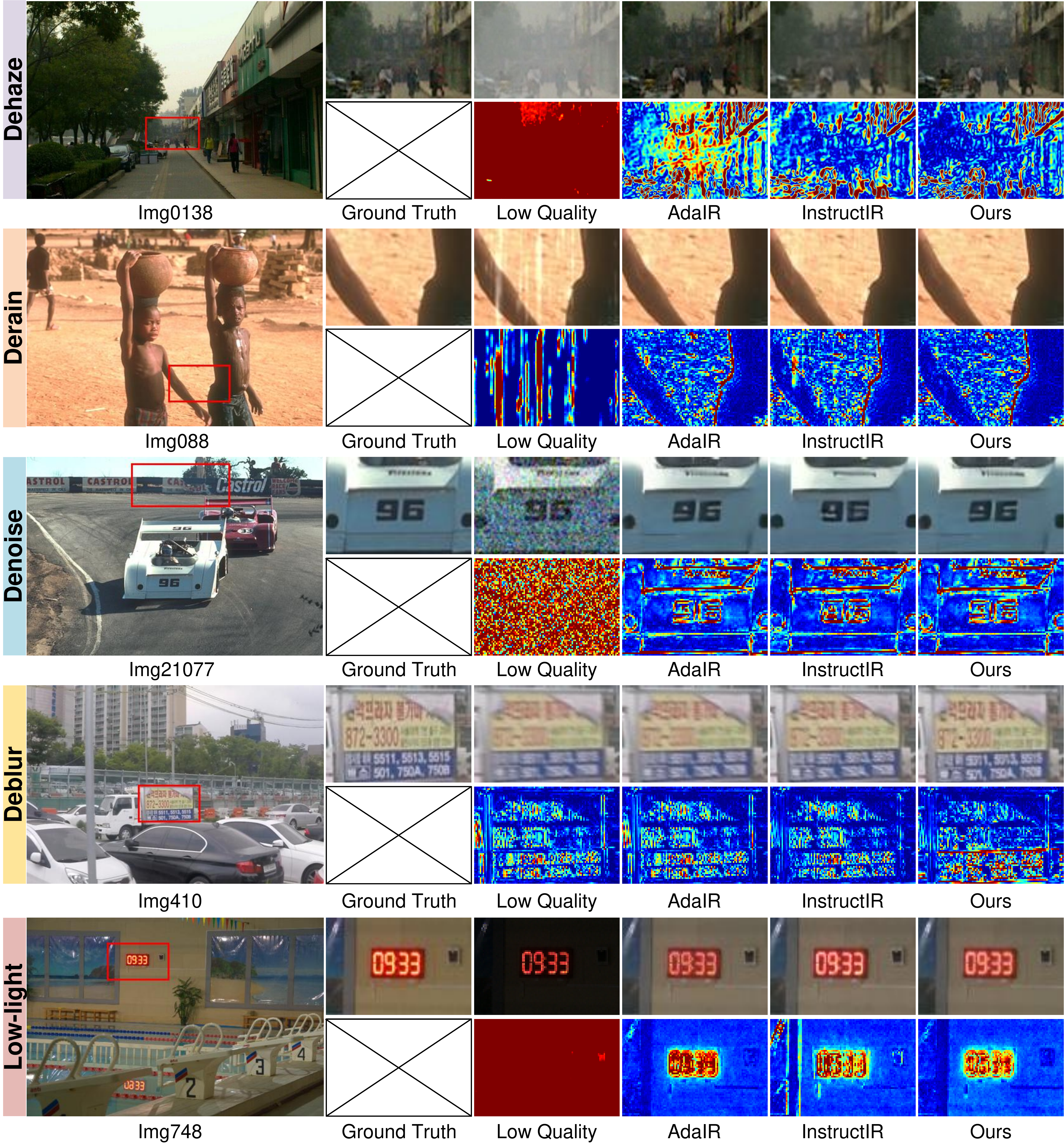}
   \caption{\major{Qualitative comparison of DRNet against state-of-the-art methods across five restoration tasks. For each task, the top row displays the magnified RGB patch, and the bottom row displays the corresponding {error map} (heatmap of absolute difference $|\boldsymbol{I}_{pred} - \boldsymbol{I}_{GT}|$).
   In the error maps, {dark blue} indicates minimal error (high fidelity), while {red/yellow} indicates high error.}}
   \vspace{-5mm}
   \label{fig:cmp_sota_quality}
\end{figure*}
\begin{table}[!t]
  \centering
  \caption{\resubmit{Performance comparison on mix-degradation dataset. The prompt used in InstructIR is ``\textit{Improve the quality of this image.}" The noise $\sigma$ is set to 25.}}
  \footnotesize 
    \resizebox{\linewidth}{!}{
    \setlength{\tabcolsep}{3pt}
    \begin{tabular}{cc  cc  cc}
         \toprule
         & & \multicolumn{2}{c}{\bf Haze+Rain} & \multicolumn{2}{c}{\bf Noise+Haze+Rain}\\
         \cmidrule(lr){3-4} \cmidrule(lr){5-6}
         \textbf{Type} & \textbf{Method} & PSNR~$\uparrow$ & SSIM~$\uparrow$ & PSNR~$\uparrow$ & SSIM~$\uparrow$\\
         \midrule
           \multirow{5}{*}{One Step} & AirNet & 18.61 & 0.776 & 14.55 & 0.607 \\
           & PromptIR & 18.97 & 0.771 & 14.53 & 0.603 \\
           & InstructIR & 14.68 & 0.663 & 14.38 & 0.439 \\
           & AdaIR & 18.68 & 0.771 & 14.55 & 0.605 \\
           & Ours w/o prompt & 19.07 & 0.774 & 14.56  & 0.610 \\
          \midrule
           \multirow{6}{*}{Multiple Steps} & AirNet & 18.44 & 0.771 & 18.58 & 0.653 \\
           & PromptIR & 18.81 & 0.813 & 18.63 & 0.669 \\
           & InstructIR & 14.57 & 0.617 & 13.84 & 0.460 \\
           & AdaIR & 18.66 & 0.810 & 18.82 & 0.668 \\
           & Ours w/o prompt & 17.85  & 0.780 & 18.95 & 0.676 \\
           & Ours & 20.34 & 0.867 & 19.61 & 0.701 \\
         \bottomrule
    \end{tabular}
    }
    \vspace{-2mm}
  \label{tab:ab_cdd}
\end{table}
\begin{figure}[!t]
  \centering
  \includegraphics[width=\linewidth]{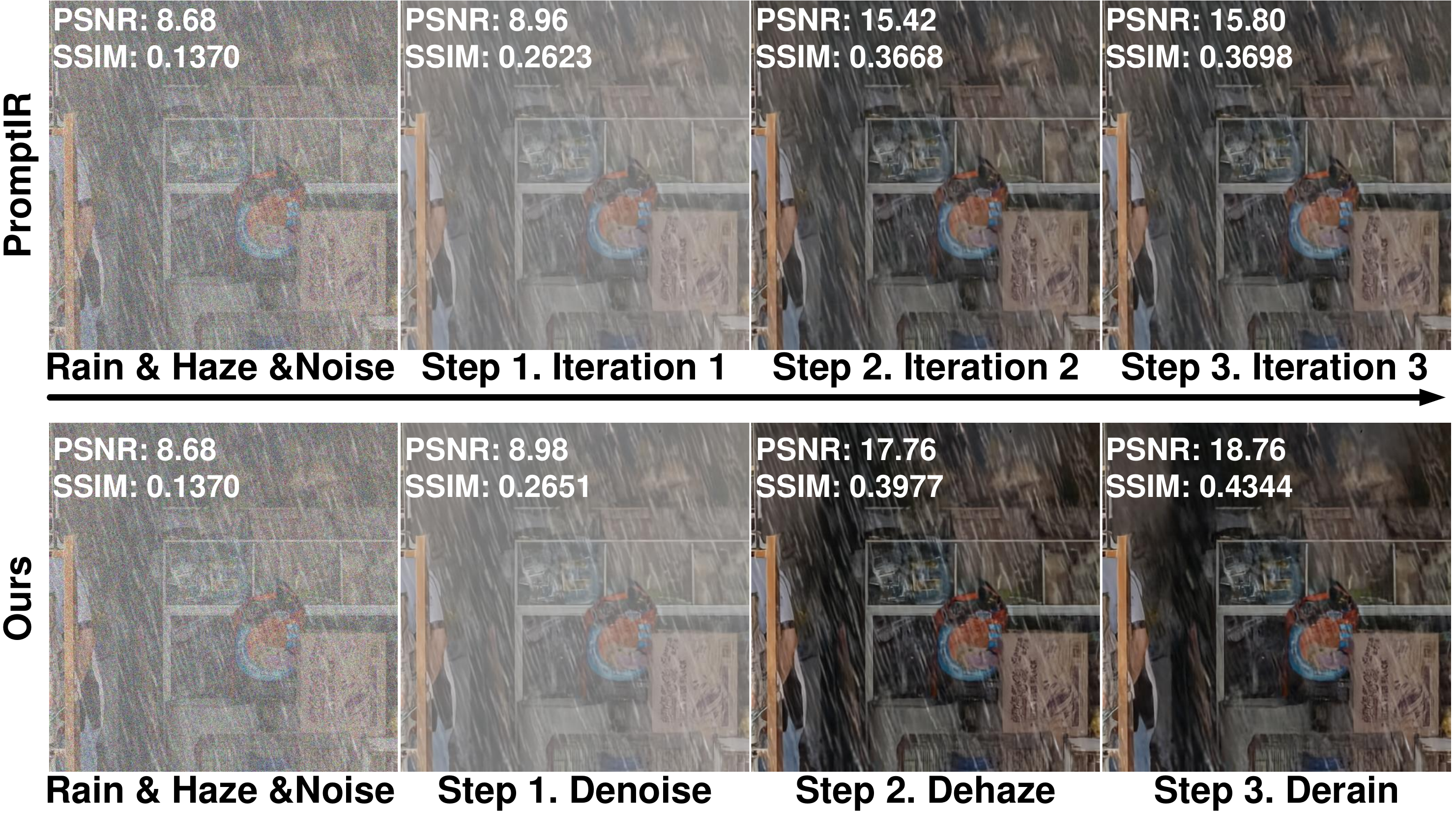}
  \caption{\major{Visual comparison of the restoration process under the mixed degradation scenario (Rain+Haze+Noise). \textbf{Top Row:} PromptIR~\cite{potlapalli2024promptir} attempts to restore the image through implicit iterations but fails to disentangle the complex corruptions, resulting in residual artifacts and poor visibility (PSNR: 15.80dB). \textbf{Bottom Row:} Our DRNet utilizes the {Sequential Specialist Strategy}. By decomposing the problem into atomic sub-tasks, our method effectively ``peels off" the degradation layers one by one. The final output recovers clear structural details, validating the effectiveness of our prior-guided decomposition approach.}}
  \vspace{-3mm}
  \label{fig:mixed_visual}
  \end{figure}
\begin{table*}[!t]
  \centering
  \caption{\resubmit{Real-world restoration results in All-in-One (\textbf{\textit{Three Tasks}} and \textbf{\textit{ Five Tasks}}) settings with state-of-the-art methods. The best results are \textbf{highlighted}.}}
  \resizebox{0.93\linewidth}{!}{
  \setlength{\tabcolsep}{3pt}
  \begin{tabular}{ccccccccccccc}
       \toprule
       & \multicolumn{6}{c}{\bf \textit{Three Tasks}} & \multicolumn{6}{c}{\bf \textit{Five Tasks}}\\
       \cmidrule(lr){2-7} \cmidrule(lr){8-13}
       & \multicolumn{2}{c}{\bf {SIDD~\cite{abdelhamed2018high}}} & \multicolumn{2}{c}{\bf {RTTS~\cite{li2018benchmarking}}} & \multicolumn{2}{c}{\bf {RealRain-1k-L~\cite{li2022toward}}} & \multicolumn{2}{c}{\bf {SIDD~\cite{abdelhamed2018high}}} & \multicolumn{2}{c}{\bf {RTTS~\cite{li2018benchmarking}}} & \multicolumn{2}{c}{\bf {RealRain-1k-L~\cite{li2022toward}}}\\
       \cmidrule(lr){2-3} \cmidrule(lr){4-5} \cmidrule(lr){6-7} \cmidrule(lr){8-9} \cmidrule(lr){10-11} \cmidrule(lr){12-13}
       \textbf{Method} & PSNR$\uparrow$ & SSIM$\uparrow$ &  FADE$\downarrow$ & NIMA$\uparrow$ & PSNR$\uparrow$ & SSIM$\uparrow$ & PSNR$\uparrow$ & SSIM$\uparrow$ & FADE$\downarrow$ & NIMA$\uparrow$ & PSNR$\uparrow$ & SSIM$\uparrow$\\
       \midrule
         AirNet~\cite{potlapalli2024promptir} & 23.86 & 0.459 & 1.534 & 4.559  & 19.88 & 0.683 & 23.64 & 0.441  & 1.627 & 4.552 &  18.64 & 0.655 \\
         PromptIR~\cite{potlapalli2024promptir} & 24.58 & 0.482 & 1.298 & 4.529 &  22.98 & 0.767 & 24.11 & 0.469 & 1.356 & 4.437 &  22.31 & 0.759\\
         InstructIR~\cite{conde2024instructir} & 24.35 & 0.479 & \bf 1.257 & 4.573 &  27.21 & 0.901 & 24.35 & 0.479  & \bf 1.257 & 4.573 &  27.21 & 0.901\\
         AdaIR~\cite{cui2025adair} & 24.85 & 0.399 & 1.475 & 4.835 &  21.96 & 0.771 & 17.58 & 0.243 & 1.498 & 4.829 &  22.70 & 0.789\\
         PerceiveIR~\cite{zhang2025perceive} & 24.88 & 0.504 & 1.213 & 4.681 &  27.79 &\bf 0.915 & 24.65 & 0.491 & 1.278 & 4.624 &  27.43 & \bf 0.903 \\
         \multicolumn{1}{c}{Ours} & \bf 28.23 & \bf 0.602 & 1.261 & \bf 4.872 &  \bf 28.10 &  0.904 & \bf 26.72 & \bf 0.758 & 1.269 &  \bf 4.883 &  \bf 27.89 & 0.899\\
       \bottomrule
  \end{tabular}
  }
  \vspace{-3mm}
  \label{tab:ab_unseen_tasks}
\end{table*}
\subsection{Comparison with State-of-the-Art Methods}
\label{sec:exp:sota_cmp}
\noindent\resubmit{\textbf{Quantitative Results.} We conducted comprehensive evaluations across five image restoration tasks: derain, dehaze, denoise, deblur, and low-light enhancement. The results are detailed in Table~\ref{tab:5tasks}. In this challenging setting, our user-guided model (``Ours") establishes a new state-of-the-art, achieving the highest average PSNR and SSIM. This superior performance is achieved with remarkable efficiency. For instance, DRNet surpasses the leading all-in-one method, AdaIR, by 0.25dB in average PSNR while being 74\% smaller, and it outperforms the powerful general restorer MambaIR with a 1.48dB average gain using 72\% fewer parameters. \major{To ensure a fair comparison, we highlight that InstructIR~\cite{conde2024instructir} also utilizes explicit user instructions. Against this direct competitor, our user-guided model demonstrates clear superiority, improving PSNR by 0.9dB while using fewer parameters. Furthermore, we present the ``Ours w/o prompt" results primarily as a baseline to validate the importance of the proposed prior-guided mechanism. While this general-purpose mode provides a competitive foundation (outperforming PromptIR~\cite{potlapalli2024promptir} which uses implicit prompts), the significant performance leap observed in the user-guided mode confirms our core thesis: blind networks face inherent limitations due to task heterogeneity, and explicitly incorporating task priors is essential to resolve these conflicts and achieve state-of-the-art restoration.}}

\major{To further analyze our model's performance in a more focused setting, we also conducted experiments on three core tasks, with results shown in Table~\ref{tab:3tasks}. Our model's superiority is reaffirmed here, as it again achieves the highest average PSNR and SSIM, matching or exceeding the top-performing baseline, AdaIR, but with significantly fewer parameters. The focused nature of this benchmark leads to slightly higher performance on shared tasks compared to the five-task setting. This suggests that while our framework effectively mitigates task interference, it remains a subtle and important factor in multi-task learning.}

\noindent\textbf{Qualitative Results.} We present a visual comparison in Figure~\ref{fig:cmp_sota_quality}, demonstrating the effectiveness of our image restoration method across five degradation tasks compared to AdaIR and InstructIR. \major{To facilitate a precise assessment of restoration fidelity, we visualize the pixel-wise error maps (absolute difference with ground truth) below each RGB result, where dark blue indicates minimal error.} In the first row, while AdaIR and InstructIR make some improvements, they fail to fully remove distant haze. \major{This is clearly reflected in their error maps, which exhibit significant high-error residuals (red/yellow patterns) in the background.} In contrast, our approach effectively restores a clearer image, \major{resulting in a predominantly deep-blue error map that indicates superior fidelity.} The second row shows that competing methods retain noticeable rain streaks, \major{visible as vertical artifacts in their corresponding error maps.} Our method, however, produces a significantly clearer result, effectively removing the rain. In the third row, other methods leave residual noise artifacts, whereas our approach restores finer details. Additionally, the comparison for deblurring illustrates that while AdaIR and InstructIR recover some legibility, our method achieves markedly sharper restorations of the Korean text. \major{Our error map demonstrates clearer structural outlines around the text characters, minimizing the blur artifacts seen in competitors.}
Lastly, for low-light enhancement, our method successfully brightens the scene and enhances the ``09:33" display. \major{The error map shows well-defined boundaries around the digits,} confirming good color accuracy and minimal noise amplification compared to other SOTA methods.

\resubmit{\subsection{Evaluations on Mixed Degradations}}
\resubmit{To rigorously test DRNet's robustness on complex, unseen corruptions, we conducted experiments on mixed-degradation scenarios, with results presented in Table~\ref{tab:ab_cdd}. We created these challenging test cases using the CDD-11 dataset~\cite{guo2024onerestore} as the base for the Haze+Rain condition. To form the more complex Noise+Haze+Rain variant, we then synthetically added Gaussian noise ($\sigma=25$) to these images. Since no models were trained on these mixtures, this serves as a challenging zero-shot generalization test.} 

\major{First, evaluating our general-purpose mode (``Ours w/o prompt") reveals its strong foundational robustness. In the ``One Step" blind inference, it achieves top-tier results, and in the ``Multiple Steps" setting, it leads the three-degradation task with 18.95dB. This confirms that even without explicit guidance, our core architecture generalizes effectively to unseen degradation combinations.}

\major{However, the full potential of our framework in handling mixed degradations is unlocked in the user-guided mode (``Ours"). By applying specialized modes in a strategic sequence (denoise$\rightarrow$dehaze$\rightarrow$derain), we achieve a groundbreaking PSNR of 19.61dB on the three-degradation task. This significant gap highlights a critical advantage: unlike blind models that treat mixed degradations as a single black-box problem, DRNet generalizes to these complex unseen scenarios by decomposing them into atomic sub-tasks. This ability to transform into a sequential specialist allows for precise, step-by-step restoration, a capability unavailable to static all-in-one models. To visually verify this, we present a step-by-step comparison in Figure~\ref{fig:mixed_visual}. As shown in the top row, blind methods like PromptIR struggle to disentangle the compounded artifacts even after multiple iterations. In contrast (bottom row), our sequential specialist approach effectively removes degradations layer-by-layer—first suppressing noise, then removing haze, and finally eliminating rain streaks. This targeted decomposition leads to a significantly cleaner output that closely matches the ground truth, as evidenced by the substantial improvements in both PSNR (+2.96dB) and SSIM metrics.}
\begin{figure*}[!t]
  \centering
   \includegraphics[width=0.92\linewidth]{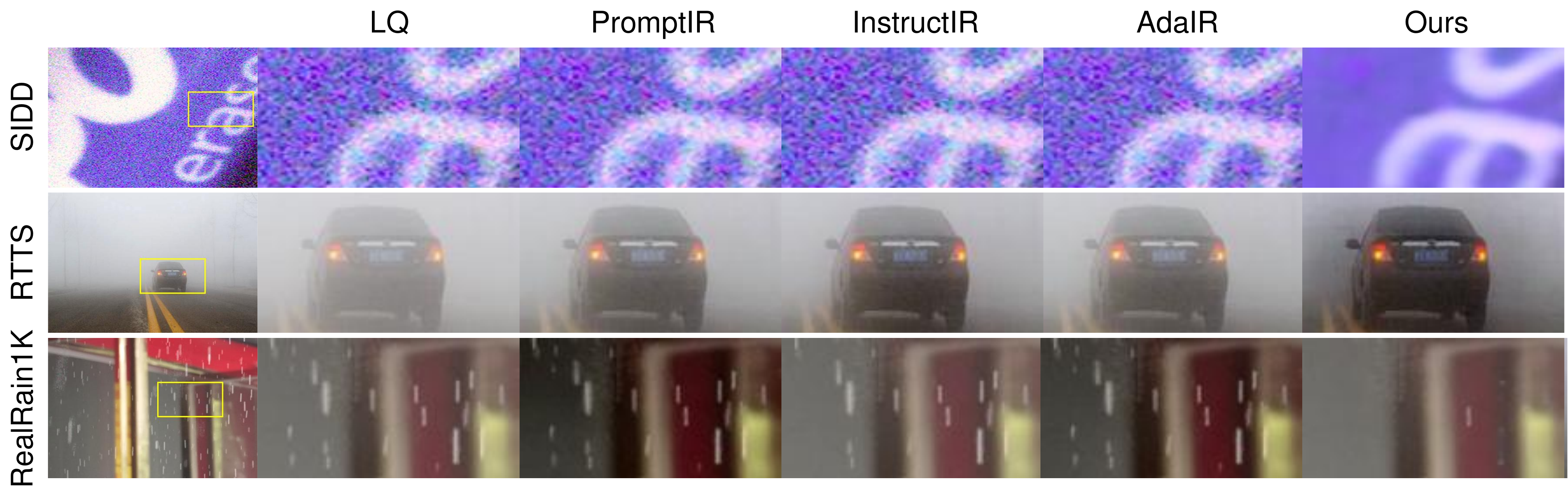}
   \caption{\resubmit{Qualitative comparison of our proposed method against existing approaches on challenging real-world degradation tasks.}}
   \vspace{-3mm}
   \label{fig:real_cmp}
\end{figure*}
\begin{table}[!t]
  \centering
  \caption{\major{Quantitative results for two settings of spatially variant degradation. The best results are \bf highlighted.}}
  \resizebox{\linewidth}{!}{
  \setlength{\tabcolsep}{12pt}
  \begin{tabular}{c | cc | cc}
       \toprule
       & \multicolumn{2}{c|}{\bf $\sigma=\{0,15,25,50\}$} & \multicolumn{2}{c}{\bf $\sigma=\{0,20,40,60\}$}\\
       \cmidrule(lr){2-3} \cmidrule(lr){4-5}
       \textbf{Method} & PSNR~$\uparrow$ & SSIM~$\uparrow$ & PSNR~$\uparrow$ & SSIM~$\uparrow$\\
       \midrule
         PromptIR & 26.72 & 0.758 & 25.69 & 0.728 \\
         InstructIR & 24.20 & 0.741 & 22.97 & 0.711 \\
         AdaIR & 25.15 & 0.746 & 23.72 & 0.717 \\
         Ours & \bf 29.06 & \bf 0.860 & \bf 27.36 & \bf 0.790 \\
       \bottomrule
  \end{tabular}
  }
  \vspace{-3mm}
  \label{tab:ab_svd_ood}
\end{table}
\resubmit{\subsection{Evaluations on Real-World Data}}
\label{ab:real}
\resubmit{To assess the practical generalization of our framework, we evaluated it on established real-world benchmarks that feature complex, non-synthetic degradations. The quantitative results are summarized in Table~\ref{tab:ab_unseen_tasks}, complemented by qualitative comparisons in Figure~\ref{fig:real_cmp}.}

\resubmit{\noindent\textbf{Quantitative Analysis.} The results demonstrate DRNet's state-of-the-art capabilities in real-world scenarios. On the SIDD dataset for real-world denoising, our method achieves a commanding performance, reaching a PSNR of 28.23dB in the three-task setting—a massive 3.35dB lead over the next-best all-in-one method, PerceiveIR. This superiority is maintained in the five-task setting, highlighting its exceptional ability to handle authentic sensor noise. On the RealRain-1k-L benchmark, our model consistently achieves the highest PSNR. Furthermore, for real-world dehazing on RTTS, DRNet obtains the highest NIMA score, indicating superior perceptual quality and visual appeal compared to all baselines. Interestingly, we observe a slight performance degradation in the five-task setting compared to the three-task one across all methods, suggesting that mitigating task interference in real-world data remains a subtle challenge.}

\resubmit{\noindent\textbf{Qualitative Results.} The visual comparisons in Figure~\ref{fig:real_cmp} corroborate our quantitative findings. On SIDD, DRNet excels at suppressing complex real-world noise while preserving fine textures, avoiding the over-smoothing artifact common in other methods. For RTTS, our model restores more vivid and natural colors while recovering sharp details in distant regions, which other models often struggle with. Finally, on RealRain-1k-L, our approach cleanly removes dense, real rain streaks with minimal residual artifacts or blurring, producing a visually superior result. In summary, both quantitative and qualitative results validate DRNet's practical effectiveness and robust generalization beyond synthetic data.}

\subsection{\major{Results on Spatially Variant Degradation}}
\major{To evaluate DRNet's local adaptability and generalization capability, we conducted experiments on spatially variant degradation, as detailed in Table~\ref{tab:ab_svd_ood}. In this setup, each test image is divided into four patches, each corrupted by a Gaussian noise level randomly selected from a specific set. We designed two rigorous settings: (1) {Seen Severities:} The noise levels $\sigma \in \{0, 15, 25, 50\}$ follow the training distribution; (2) {Unseen Severities:} The noise levels $\sigma \in \{0, 20, 40, 60\}$ are entirely unseen during training.}
\begin{table}[!t]
  \centering
  \caption{Performance comparison of general image restoration and all-in-one methods on image denoising. The best and second-best results for each setting are \textcolor{red}{\textbf{highlighted}} and \textcolor{blue}{\underline{underlined}}, respectively.}
  \resizebox{\linewidth}{!}{
  \setlength{\tabcolsep}{4pt}
  \begin{tabular}{l | ccc | ccc | ccc}
       \toprule
       &      \multicolumn{3}{c|}{\bf BSD68~\cite{martin2001database}} & \multicolumn{3}{c|}{\bf Urban100~\cite{huang2015single}} & \multicolumn{3}{c}{\bf Kodak24~\cite{franzen1999kodak}}\\
       \cmidrule(lr){2-4} \cmidrule(lr){5-7} \cmidrule(lr){8-10}
       \textbf{Method} & \textbf{15} & \textbf{25} & \textbf{50} & \textbf{15} & \textbf{25} & \textbf{50} & \textbf{15} & \textbf{25} & \textbf{50} \\
       \midrule
       IRCNN~\cite{zhang2017learning}   & 33.86 & 31.16 & 27.86 & 33.78 & 31.20 & 27.70 & 34.69 & 32.18 & 28.93 \\

       FFDNet~\cite{FFDNetPlus}  & 33.87 & 31.21 & 27.96 & 33.83 & 31.40 & 28.05 & 34.63 & 32.13 & 28.98 \\
       
       DnCNN~\cite{DnCNN}  & 33.90 & 31.24 & 27.95 & 32.98 & 30.81 & 27.59 & 34.60 & 32.14 & 28.95 \\
       MPRNet~\cite{zamir2021multi} & 34.01 & 31.35 & 28.08 & \textcolor{blue}{\underline{34.13}} & \textcolor{blue}{\underline{31.75}} & \textcolor{blue}{\underline{28.41}} & 34.77 & 32.31 & 29.11 \\
       NAFNet~\cite{chen2022simple}     & 33.67 & 31.02 & 27.73 & 33.14 & 30.64 & 27.20 & 34.27 & 31.80 & 28.62 \\
       HINet~\cite{chen2021hinet}      & 33.72 & 31.00 & 27.63 & 33.49 & 30.94 & 27.32 & 34.38 & 31.84 & 28.52 \\
       DGUNet~\cite{mou2022deep}     & 33.85 & 31.10 & 27.92 & 33.67 & 31.27 & 27.94 & 34.56 & 32.10 & 28.91 \\
       MIRNetV2~\cite{zamir2020learning}   & 33.66 & 30.97 & 27.66 & 33.30 & 30.75 & 27.22 & 34.29 & 31.81 & 28.55 \\
       SwinIR~\cite{liang2021swinir}     & 33.31 & 30.59 & 27.13 & 32.79 & 30.18 & 26.52 & 33.89 & 31.32 & 27.93 \\
       Restormer~\cite{zamir2022restormer}  & 34.03 & 31.49 & 28.11 & 33.72 & 31.26 & 28.03 & 34.78 & 32.37 & 29.08 \\
       \midrule
       \midrule
         DL~\cite{fan2019general}             & 23.16 & 23.09 & 22.09 & 21.10 & 21.28 & 20.42 & 22.63 & 22.66 & 21.95 \\
         T.weather~\cite{valanarasu2022transweather}   & 31.16 & 29.00 & 26.08 & 29.64 & 27.97 & 26.08 & 31.67 & 29.64 & 26.74 \\
         TAPE~\cite{liu2022tape}           & 32.86 & 30.18 & 26.63 & 32.19 & 29.65 & 25.87 & 33.24 & 30.70 & 27.19 \\
         AirNet~\cite{li2022all}         & 33.49 & 30.91 & 27.66 & 33.16 & 30.83 & 27.45 & 34.14 & 31.74 & 28.59 \\
         IDR~\cite{zhang2023ingredient}            & \textcolor{blue}{\underline{34.11}} & \textcolor{red}{\bf 31.60} & 28.14 & 33.82 & 31.29 & 28.07 & 34.78 & 32.42 & 29.13 \\
         InstructIR~\cite{conde2024instructir}        & 34.00 & 31.40 & \textcolor{blue}{\underline{28.15}} & 33.77 & 31.40 & 28.13 & 34.70 & 32.26 & 29.16 \\
         AdaIR~\cite{cui2025adair} & 34.01 & 31.35 & 28.06 & 34.10 & 31.68 & 28.29 & \textcolor{blue}{\underline{34.89}} & \textcolor{blue}{\underline{32.38}} & \textcolor{blue}{\underline{29.21}} \\
         Ours & \textcolor{red}{\bf 34.19} & \textcolor{blue}{\underline{31.54}} & \textcolor{red}{\bf 28.26} & \textcolor{red}{\bf 34.45} & \textcolor{red}{\bf 32.15} & \textcolor{red}{\bf 28.89} & \textcolor{red}{\bf 35.17} & \textcolor{red}{\bf 32.71} & \textcolor{red}{\bf 29.56} \\
       \bottomrule
  \end{tabular}
  }
  \vspace{-4mm}
  \label{tab:ab_ood}
\end{table}
\begin{figure*}[!t]
  \centering
   \begin{overpic}[scale=.243]{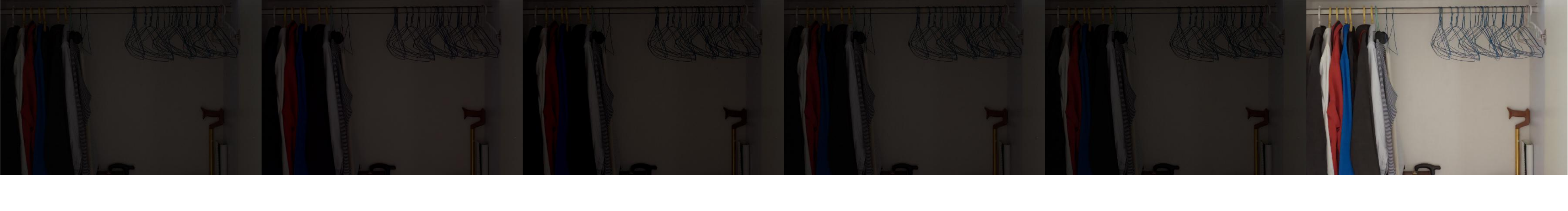}
    \put(4.5,0.5){\footnotesize Low Light}
    \put(20.5,0.5){\footnotesize $\phi(\textit{``denoise"})$}
    \put(37,0.5){\footnotesize $\phi(\textit{``dehaze"})$}
    \put(54,0.5){\footnotesize $\phi(\textit{``derain"})$}
    \put(70.8,0.5){\footnotesize $\phi(\textit{``deblur"})$}
    \put(86.7,0.5){\footnotesize $\phi(\textit{``enhance"})$}
   \end{overpic}
   \caption{\major{Effectiveness of Prompt. The model restores the image only when given the correct prompt. Incorrect prompts such as \textit{``denoise"}, \textit{``dehaze"}, \textit{``derain"}, \textit{``deblur"} fail to produce the desired restoration, while the correct prompt \textit{``enhance"} successfully restores the image.}}
   \label{fig:ab_prompts}
\end{figure*}
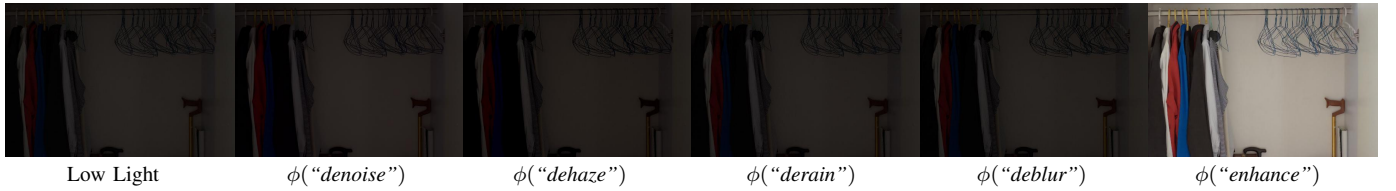

\major{As shown in the table, DRNet demonstrates exceptional robustness. Under the first setting, it outperforms the second-best method by {2.34dB} in PSNR. More importantly, in the second setting with unseen noise levels, our model maintains a substantial lead of {1.67dB}. These results empirically confirm two key properties of our framework: (1) The TSM sets a global ``functional domain" (denoising), while the backbone effectively handles spatially variant severities without requiring pixel-wise prompt tuning; (2) The model learns a generalized restoration policy that extends well beyond the discrete noise levels seen during training.}
\subsection{Comparison with Task-specific Methods}
To validate the upper bounds of our framework's specialization capabilities, we conducted a comprehensive evaluation on the challenging image denoising task. The results, presented in Table~\ref{tab:ab_ood}, compare DRNet not only against other all-in-one models but also against state-of-the-art methods designed exclusively for denoising. The results reveal a remarkable finding: our unified all-in-one model consistently achieves state-of-the-art performance, matching or even surpassing the top specialist denoisers. This is particularly evident on the complex Urban100 and Kodak24 datasets. For instance, on Urban100, which features intricate textures, our model outperforms powerful specialist methods like MPRNet and Restormer across all noise levels. This demonstrates that our prior-guided reconfiguration can create an inference-time network that is as good as, or better than, a model statically designed and trained for this single task. Furthermore, when compared to its direct all-in-one competitors, DRNet's superiority is even more pronounced. Against the leading baseline, AdaIR, our model achieves significant PSNR gains of up to 0.60dB on Urban100 and 0.35dB on Kodak24. The advantage over InstructIR is even larger, with improvements reaching 0.76dB on Urban100. This state-of-the-art performance in a specialized task underscores the power of our dynamic reparameterization paradigm, proving that a single, efficient framework can indeed specialize to an expert level that obviates the need for separate, task-specific models.
\begin{table}[!t]
  \centering
  \caption{\resubmit{Comparison of model complexity across different methods.}}
  \footnotesize 
   \setlength{\tabcolsep}{5pt}
  \begin{tabular}{cccccc}
    \toprule
     & Params & FLOPs & Max Mem. & PSNR & Run Time\\
     Method & (M) & (G) & (M) & (dB) & (ms) \\
    \midrule
    PromptIR  & 32.96 & 43.17 & 3538.59 & 29.15 & 39.35\\ 
    AdaIR & 28.77 & 40.44 & 3479.30 & 30.20 & 46.68\\  
    Ours w/o Init. & 16.49 & 34.59 & 1633.98 & 30.45 & 47.78\\
    Ours & 7.39 & 16.48 & 911.58 & 30.45 & 28.92 \\
    \bottomrule
  \end{tabular}
  \label{tab:ab_complexity}
\end{table}
\resubmit{\subsection{Complexity and Runtime Analysis}}
\label{exp:complexity}
\resubmit{To substantiate the practicality of our proposed model, we provide a detailed computational analysis in Table~\ref{tab:ab_complexity}, comparing DRNet against state-of-the-art methods on a 128×128 image using a single NVIDIA RTX 5090 GPU. The results highlight DRNet's exceptional efficiency. With only 7.39M parameters, it is over 74\% smaller than AdaIR and 77\% smaller than PromptIR, leading to a memory footprint of just 911.58 MB—less than 26\% of that required by competitors. Crucially, this dramatic reduction in complexity does not come at the cost of performance. DRNet achieves a PSNR of 30.47 dB, significantly outperforming PromptIR and even surpassing the larger AdaIR model. This superior computational efficiency, with only 16.48 GFLOPs, translates directly into a significant speed advantage: DRNet's average runtime of 28.92 ms is approximately 38\% faster than AdaIR and 26\% faster than PromptIR, making it ideal for practical, real-time applications.}

\resubmit{The core of DRNet's efficiency lies in its dynamic reparameterization mechanism. This is quantitatively proven by the ``Ours w/o Init." ablation study, where the multi-branch training architecture is used for inference without fusion. In this state, FLOPs more than double to 34.59G and latency increases by 65\% to 47.78 ms, while the PSNR remains identical. This comparison provides definitive proof that our reparameterization strategy slashes the computational burden by over 52\% without any performance loss, allowing DRNet to harness the representational power of a complex architecture during training while deploying as a streamlined, ultra-efficient network for inference.}
\begin{table}[!t]
  \centering
  \caption{\major{Ablation analysis of component contributions to model efficiency and performance. ``Train/Inf." denotes the statistics during Training and Inference stages. Note that ``w/o CWTE" implies replacing the wavelet encoder with a conventional encoder composed of stacked DRBs, highlighting the efficiency of the wavelet design.}}
  \setlength{\tabcolsep}{6pt}
  \begin{tabular}{lccc}
       \toprule
        &   Params (M) & FLOPs (G) & Average\\
       Variant & (Train / Inf.) & (Train / Inf.) & (PSNR / SSIM) \\
       \midrule
      \rowcolor{gray!10} Full DRNet  & 16.49 / 7.39 & 34.59 / 16.48 & 30.24 / 0.911 \\
      \midrule
      \quad w/o CWTE  & 39.27 / 17.96 & 35.28 / 17.16 & 29.72 / 0.899 \\
      \quad w/o DRMLP & 7.39 / 7.39 & 16.48 / 16.48 & 29.89 / 0.903 \\
      \quad w/o TSM  & 16.29 / 7.19 & 34.59 / 16.48 & 29.17 / 0.897 \\
       \bottomrule
  \end{tabular}
  \label{tab:ab_dwt_tgfm}
\end{table}
\subsection{Ablation Studies}
\label{sec:exp:ablation}
\subsubsection{\major{Effectiveness of Prompt}} 
\major{Since we use task-specific prompts to control the restoration process, it is crucial to evaluate the model's performance when given an incorrect prompt. The results are illustrated in Figure~\ref{fig:ab_prompts}. Under low-light conditions, we tested different prompts such as \textit{``denoise"}, \textit{``dehaze"}, \textit{``derain"} and \textit{``deblur"}, and visualized the corresponding restoration outcomes. The results demonstrate that when the prompt does not align with the actual degradation, the model fails to produce an appropriate response. However, when the correct prompt, \textit{``enhance"}, is provided, the model successfully generates the desired restoration output.}
\subsubsection{Effectiveness of CWTE}
\major{We evaluate the effectiveness of the CWTE in Table~\ref{tab:ab_dwt_tgfm}. When replacing the CWTE with a conventional symmetric encoder composed of stacked DRBs (``w/o CWTE"), the model suffers from a dual disadvantage: a performance drop of 0.52dB and a massive increase in inference parameters from 7.39M to 17.96M (+143\%). This quantitative evidence confirms that CWTE is the cornerstone of our model's lightweight design, enabling efficient feature extraction without the burden of heavy convolutional stacks.}

\begin{figure}[!t]
  \centering
  \includegraphics[width=\linewidth]{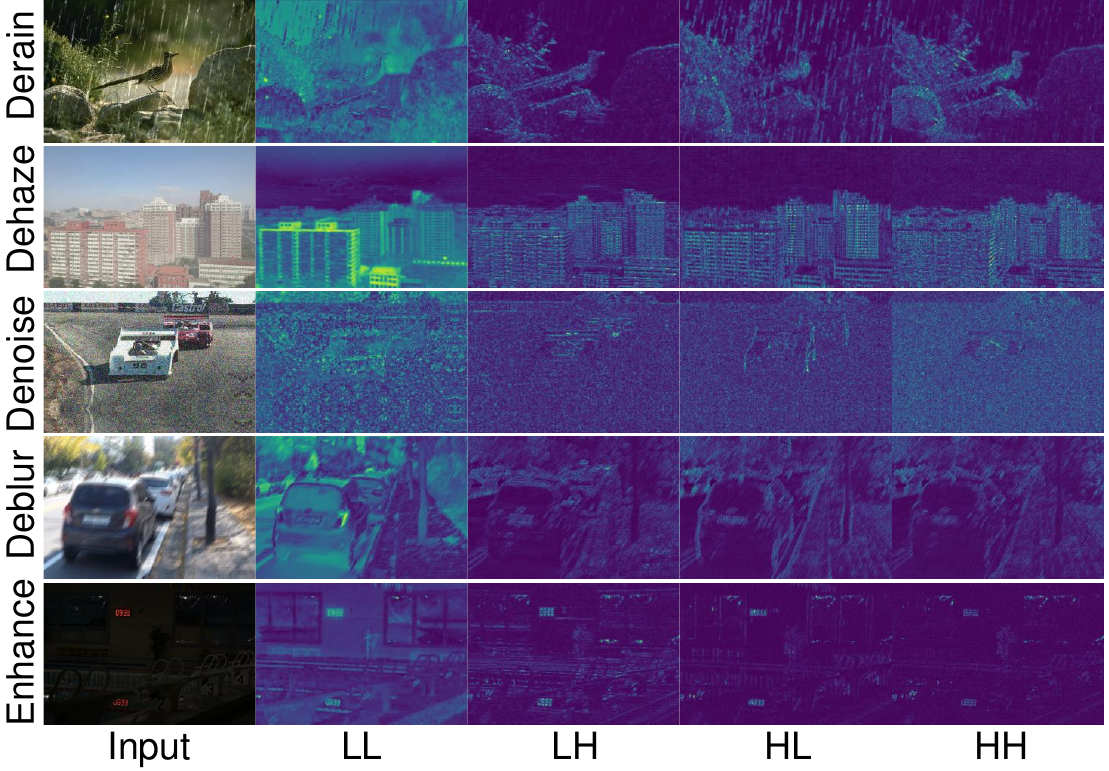}
  \caption{\resubmit{Visualization of decomposed intermediate features within our CWTE module across five different restoration tasks. The figure demonstrates that our model learns to isolate distinct degradation patterns into specific frequency sub-bands.}}
  \label{fig:ab_dwt}
  \vspace{-3mm}
\end{figure}

\resubmit{Additionally, to provide a deeper qualitative understanding of CWTE's effectiveness, we visualize the decomposed intermediate features at the first level of our encoder, as illustrated in Figure~\ref{fig:ab_dwt}. The visualization across five diverse tasks reveals that our model has learned to intelligently disentangle degradation artifacts from image content in the frequency domain. For deraining and denoising, high-frequency artifacts like rain streaks and noise are successfully isolated into the LH, HL, and HH sub-bands. For dehazing, the model addresses the low-frequency nature of haze in the LL band while preserving structural details in the high-frequency bands. Similarly, for low-light enhancement, the LL band captures the primary brightness information, while for deblurring, the model focuses on restoring the lost high-frequency details. This compelling evidence demonstrates that CWTE enables the network to perform targeted, task-specific manipulations in the frequency domain, which is crucial for its strong performance and efficiency.}

\subsubsection{Effectiveness of DRMLP}
\major{We evaluate the impact of our reparameterization strategy. The ``w/o DRMLP" variant in Table~\ref{tab:ab_dwt_tgfm} represents a static model trained with a single branch (setting bank size to 1). The results highlight the advantage of our approach: although both models share the exact same inference cost (7.39M parameters), our full DRNet utilizes a multi-branch structure (16.49M parameters) during training to learn richer features. Collapsing these branches for inference allows us to achieve a 0.35dB performance gain over the static counterpart with zero extra inference overhead. This confirms that the parallel linear design is essential for enhancing representation power within a compact inference architecture.}

\subsubsection{Effectiveness of TSM}
\major{To address the critical challenge of degradation heterogeneity, we introduce the Task-Specific Modulator (TSM). As shown in Table~\ref{tab:ab_dwt_tgfm}, removing the TSM (``w/o TSM") results in the most significant performance degradation among all variants, with average PSNR dropping by 1.07dB. Notably, this component introduces negligible parameter overhead (approx. 0.2M difference). The drastic performance drop upon its removal strongly suggests that without the TSM's dynamic weight modulation, the shared network struggles to optimize conflicting objectives across heterogeneous tasks. Thus, TSM plays a decisive role in resolving task interference with minimal computational cost.}
\begin{figure}[!t]
  \centering
   \includegraphics[width=\linewidth]{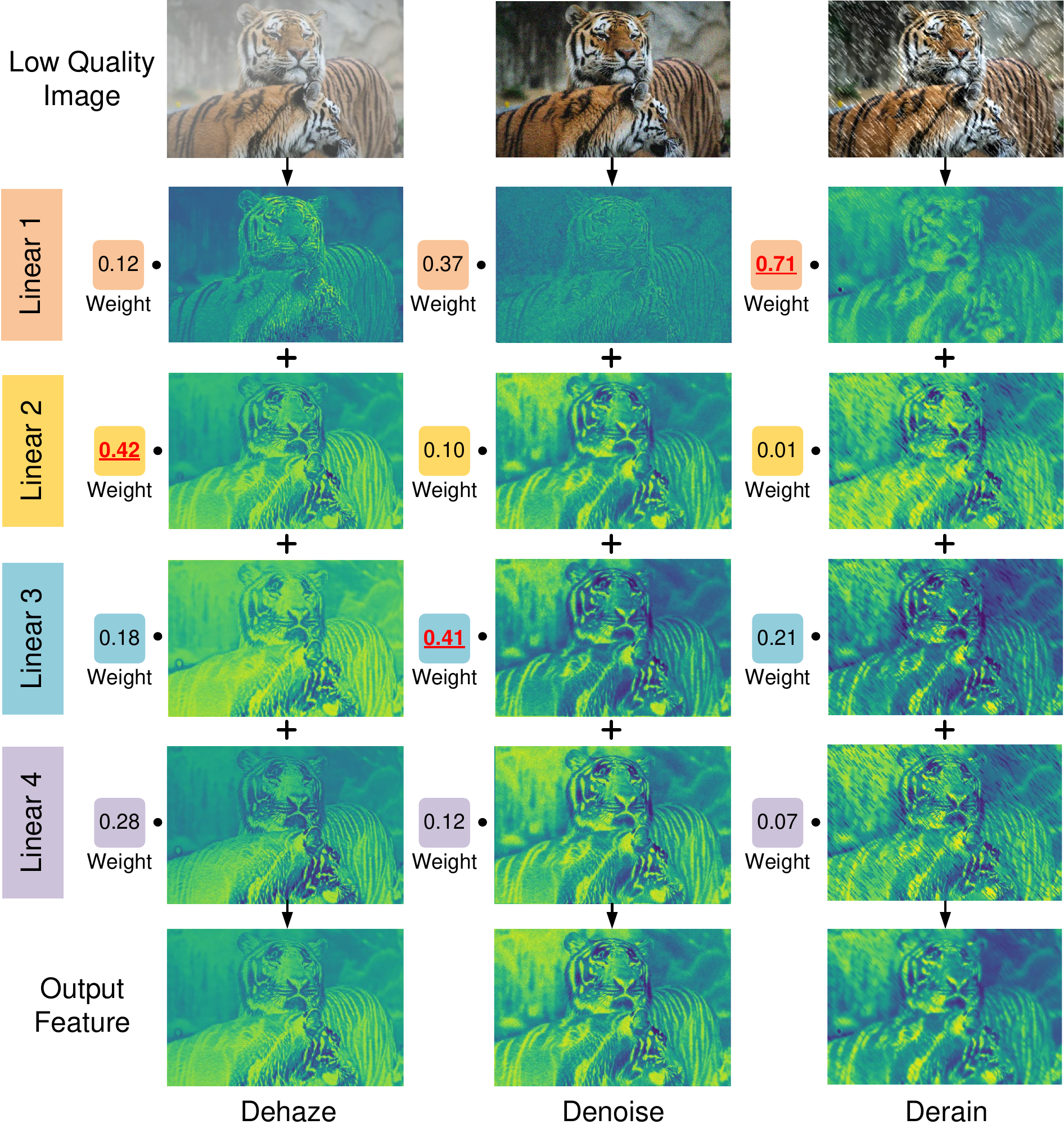}
   \caption{\resubmit{Visualization of TSM's dynamic feature modulation. For three distinct tasks (Dehaze, Denoise, Derain), our Task-Specific Modulator (TSM) assigns unique weights to the four parallel linear branches. This leads to visibly different intermediate feature maps and a final output feature that is more pronounced and task-relevant.}}
   \vspace{-5mm}
   \label{fig:ab_ctrl_fea}
\end{figure}
\begin{figure}[t]
  \centering
   \includegraphics[width=\linewidth]{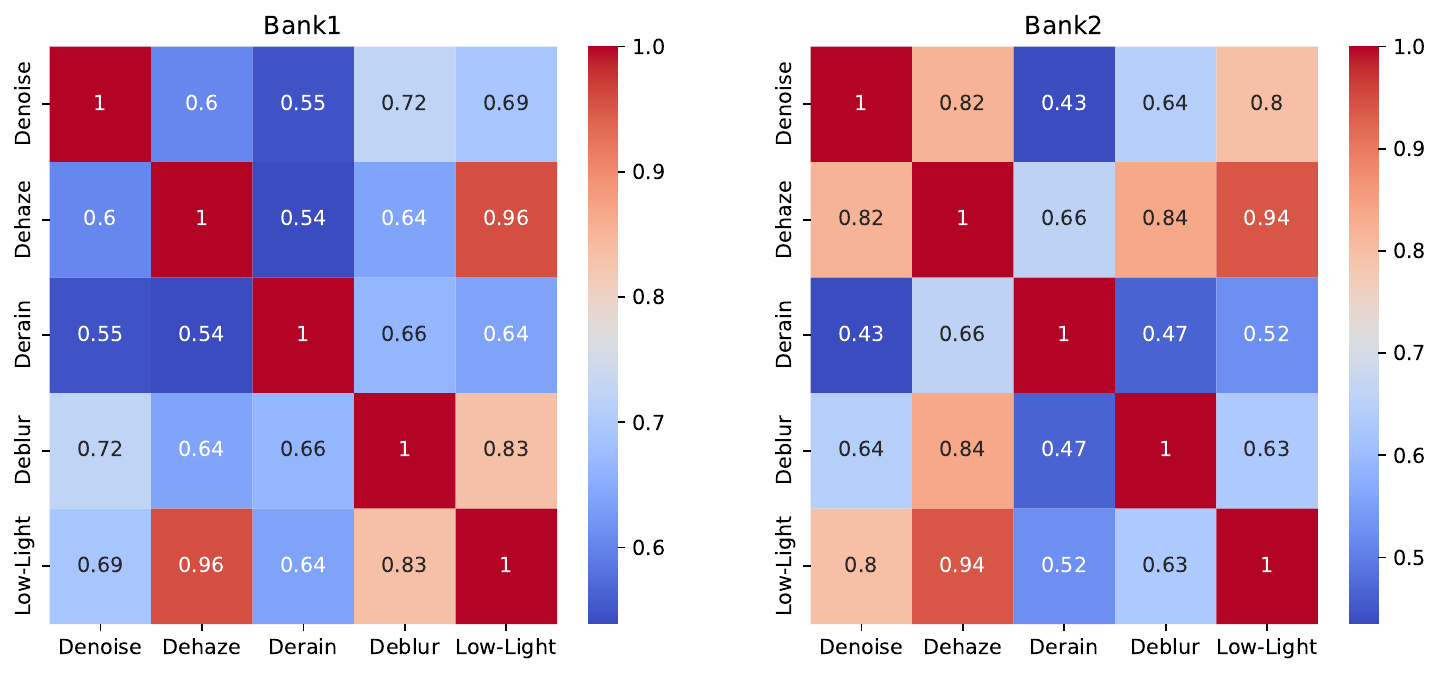}
   \caption{\resubmit{Visualization of TSM-induced parameter space specialization. The heatmaps show the cosine similarity of the final, fused DRMLP weights for five tasks. Two key behaviors are revealed: (1) Structurally dissimilar tasks like denoise and derain show low similarity (\eg, 0.55), indicating decoupled parameter spaces. (2) Semantically related tasks like dehaze and low-light show high similarity (\eg, 0.94), indicating shared knowledge.}}
   \vspace{-5mm}
   \label{fig:tsm_weight_similarity}
\end{figure}
\subsubsection{\resubmit{Interpretability of TSM}}
\label{ab:analysis_TSM}
\resubmit{To provide deeper insight into our framework's mechanism, we analyze the behavior of the TSM at two key levels: feature-level adaptation within a block and parameter-space specialization across tasks.}

\resubmit{\noindent\textbf{Feature-Level Adaptation.} Figure~\ref{fig:ab_ctrl_fea} visualizes the feature maps from the parallel branches of a DRMLP block for three distinct tasks. The figure provides direct evidence of the TSM's intelligent, task-driven modulation. For each task, the TSM assigns a unique set of weights to the four parallel branches. This dynamic weighting leads to visibly different intermediate feature activations; for instance, the denoise task heavily relies on the transformation from Linear 3 (weight 0.41), resulting in a distinct activation pattern. This confirms that the TSM effectively guides the network to learn and activate the most appropriate feature transformations for the specific task at hand.}

\resubmit{\noindent\textbf{Parameter-Space Specialization.} At a higher level, Figure~\ref{fig:tsm_weight_similarity} reveals the TSM's sophisticated strategy for managing a multi-task environment by visualizing the cosine similarity of the final, fused DRMLP weights across five tasks. The heatmaps show that structurally dissimilar tasks, such as denoise and derain, exhibit low weight similarity (\eg, 0.55 in Bank1). This indicates that the TSM creates largely decoupled, specialized parameter spaces to mitigate negative task interference. Conversely, semantically related tasks, like dehazing and low-light enhancement (both involving global visibility improvement), yield configurations with very high similarity (0.94 in Bank2). This demonstrates that the TSM also intelligently facilitates knowledge sharing between related tasks. In summary, these visualizations prove that the TSM operates as an intelligent controller, performing both fine-grained feature modulation and coarse-grained parameter-space management across the entire task suite.}

\section{Conclusion}
\label{sec:conclusion}
\major{In this paper, we presented the Dynamic Reparameterization Network (DRNet), a novel framework that resolves the dilemma between performance and efficiency in all-in-one image restoration. The core innovation lies in our {Prior-Guided Network Reconfiguration} paradigm. Unlike existing methods that rely on heavy, per-input degradation estimation, DRNet utilizes the {Task-Specific Modulator (TSM)} to translate abstract task priors into specific architectural weights at the initialization stage. This mechanism not only ensures {zero computational overhead} during inference but also effectively mitigates the challenge of {task interference} by disentangling conflicting restoration goals into specialized modes. Furthermore, by incorporating the Continuous Wavelet Transform Encoder (CWTE), we explicitly leverage frequency domain characteristics to achieve a lightweight yet powerful representation. Extensive experiments demonstrate that DRNet establishes a new state-of-the-art across five restoration tasks. Crucially, it showcases unique flexibility, excelling as both a robust general-purpose model in blind scenarios and a top-performing specialist when guided by user priors. Our current framework assumes atomic task priors. While compound degradations can be effectively addressed via the proposed sequential specialist application strategy, directly supporting compositional prompts (e.g., ``denoise and dehaze") within a single pass remains an interesting avenue for future research.}

\ifCLASSOPTIONcaptionsoff
  \newpage
\fi

\bibliographystyle{IEEEtran}
\bibliography{IEEEabrv,reference}

@String(TPAMI = {IEEE Trans. Pattern Anal. Mach. Intell.})

@String(IJCV = {Int. J. Comput. Vis.})

@String(CVPR= {IEEE Conf. Comput. Vis. Pattern Recog.})

@String(ICCV= {Int. Conf. Comput. Vis.})

@String(ECCV= {Eur. Conf. Comput. Vis.})

@String(NIPS= {Adv. Neural Inform. Process. Syst.})

@String(BMVC= {Brit. Mach. Vis. Conf.})

@String(TIP  = {IEEE Trans. Image Process.})

@String(TMM  = {IEEE Trans. Multimedia})

@String(ACMMM= {ACM Int. Conf. Multimedia})

@String(ICLR = {Int. Conf. Learn. Represent.})

@String(AAAI = {AAAI})

@String(ICCVW= {Int. Conf. Comput. Vis. Worksh.})

@String(ICML = {ICML})

@String(TPAMI  = {IEEE TPAMI})

@String(IJCV  = {IJCV})

@String(CVPR  = {CVPR})

@String(ICCV  = {ICCV})

@String(ECCV  = {ECCV})

@String(NIPS  = {NeurIPS})

@String(BMVC  =	{BMVC})

@String(TIP   = {IEEE TIP})

@String(TMM   =	{IEEE TMM})

@String(ACMMM = {ACM MM})

@String(ICLR  = {ICLR})

@String(ICCVW= {ICCVW})

@inproceedings{Liu2021iccv,
author={Liu, Ze and Lin, Yutong and Cao, Yue and Hu, Han and Wei, Yixuan and Zhang, Zheng and Lin, Stephen and Guo, Baining},
booktitle = ICCV,
pages = {9992--10002},
title = {Swin Transformer: Hierarchical Vision Transformer using Shifted Windows},
year = {2021}
}

@inproceedings{zamir2022restormer,
  title={Restormer: Efficient transformer for high-resolution image restoration},
  author={Zamir, Syed Waqas and Arora, Aditya and Khan, Salman and Hayat, Munawar and Khan, Fahad Shahbaz and Yang, Ming-Hsuan},
  booktitle=CVPR,
  pages={5728--5739},
  year={2022}
}

@ARTICLE{wen_allinone_tmm25,
  author={Wen, Yuanbo and Gao, Tao and Li, Ziqi and Zhang, Jing and Zhang, Kaihao and Chen, Ting},
  journal=TMM, 
  title={All-in-one Weather-degraded Image Restoration via Adaptive Degradation-aware Self-prompting Model}, 
  year={2025},
  volume={},
  number={},
  pages={1-13}
}

@inproceedings{guo2024onerestore,
  title={Onerestore: A universal restoration framework for composite degradation},
  author={Guo, Yu and Gao, Yuan and Lu, Yuxu and Zhu, Huilin and Liu, Ryan Wen and He, Shengfeng},
  booktitle=ECCV,
  pages={255--272},
  year={2024},
  organization={Springer}
}

@inproceedings{yan2025mobileie,
  title={MobileIE: An Extremely Lightweight and Effective ConvNet for Real-Time Image Enhancement on Mobile Devices},
  author={Yan, Hailong and Li, Ao and Zhang, Xiangtao and Liu, Zhe and Shi, Zenglin and Zhu, Ce and Zhang, Le},
  booktitle=ICCV,
  year={2025}
}

@inproceedings{li2022all,
  title={All-in-one image restoration for unknown corruption},
  author={Li, Boyun and Liu, Xiao and Hu, Peng and Wu, Zhongqin and Lv, Jiancheng and Peng, Xi},
  booktitle=CVPR,
  pages={17452--17462},
  year={2022}
}

@article{potlapalli2024promptir,
  title={Promptir: Prompting for all-in-one image restoration},
  author={Potlapalli, Vaishnav and Zamir, Syed Waqas and Khan, Salman H and Shahbaz Khan, Fahad},
  journal=NIPS,
  year={2023}
}

@article{arbelaez2010contour,
  title={Contour detection and hierarchical image segmentation},
  author={Arbelaez, Pablo and Maire, Michael and Fowlkes, Charless and Malik, Jitendra},
  journal=TPAMI,
  volume={33},
  number={5},
  pages={898--916},
  year={2010}
}

@article{ma2016waterloo,
  title={Waterloo exploration database: New challenges for image quality assessment models},
  author={Ma, Kede and Duanmu, Zhengfang and Wu, Qingbo and Wang, Zhou and Yong, Hongwei and Li, Hongliang and Zhang, Lei},
  journal=TIP,
  volume={26},
  number={2},
  pages={1004--1016},
  year={2016}
}

@inproceedings{martin2001database,
  title={A database of human segmented natural images and its application to evaluating segmentation algorithms and measuring ecological statistics},
  author={Martin, David and Fowlkes, Charless and Tal, Doron and Malik, Jitendra},
  booktitle=ICCV,
  volume={2},
  pages={416--423},
  year={2001}
}

@inproceedings{huang2015single,
  title={Single image super-resolution from transformed self-exemplars},
  author={Huang, Jia-Bin and Singh, Abhishek and Ahuja, Narendra},
  booktitle=CVPR,
  pages={5197--5206},
  year={2015}
}

@inproceedings{yang2017deep,
  title={Deep joint rain detection and removal from a single image},
  author={Yang, Wenhan and Tan, Robby T and Feng, Jiashi and Liu, Jiaying and Guo, Zongming and Yan, Shuicheng},
  booktitle=CVPR,
  pages={1357--1366},
  year={2017}
}

@article{li2018benchmarking,
  title={Benchmarking single-image dehazing and beyond},
  author={Li, Boyi and Ren, Wenqi and Fu, Dengpan and Tao, Dacheng and Feng, Dan and Zeng, Wenjun and Wang, Zhangyang},
  journal=TIP,
  volume={28},
  number={1},
  pages={492--505},
  year={2018}
}

@inproceedings{cui2024omni,
  title={Omni-kernel network for image restoration},
  author={Cui, Yuning and Ren, Wenqi and Knoll, Alois},
  booktitle=AAAI,
  volume={38},
  number={2},
  pages={1426--1434},
  year={2024}
}

@inproceedings{li2023efficient,
  title={Efficient and explicit modelling of image hierarchies for image restoration},
  author={Li, Yawei and Fan, Yuchen and Xiang, Xiaoyu and Demandolx, Denis and Ranjan, Rakesh and Timofte, Radu and Van Gool, Luc},
  booktitle=CVPR,
  pages={18278--18289},
  year={2023}
}

@article{wu2024one,
  title={One-step effective diffusion network for real-world image super-resolution},
  author={Wu, Rongyuan and Sun, Lingchen and Ma, Zhiyuan and Zhang, Lei},
  journal=NIPS,
  volume={37},
  pages={92529--92553},
  year={2024}
}

@article{li2022toward,
  title={Toward real-world single image deraining: A new benchmark and beyond},
  author={Li, Wei and Zhang, Qiming and Zhang, Jing and Huang, Zhen and Tian, Xinmei and Tao, Dacheng},
  journal={arXiv preprint arXiv:2206.05514},
  year={2022}
}

@inproceedings{abdelhamed2018high,
  title={A high-quality denoising dataset for smartphone cameras},
  author={Abdelhamed, Abdelrahman and Lin, Stephen and Brown, Michael S},
  booktitle=CVPR,
  pages={1692--1700},
  year={2018}
}

@inproceedings{li2024light,
  title={Light the night: A multi-condition diffusion framework for unpaired low-light enhancement in autonomous driving},
  author={Li, Jinlong and Li, Baolu and Tu, Zhengzhong and Liu, Xinyu and Guo, Qing and Juefei-Xu, Felix and Xu, Runsheng and Yu, Hongkai},
  booktitle=CVPR,
  pages={15205--15215},
  year={2024}
}

@inproceedings{tu2022maxim,
  title={Maxim: Multi-axis mlp for image processing},
  author={Tu, Zhengzhong and Talebi, Hossein and Zhang, Han and Yang, Feng and Milanfar, Peyman and Bovik, Alan and Li, Yinxiao},
  booktitle=CVPR,
  pages={5769--5780},
  year={2022}
}

@article{jiang2024survey,
  title={A survey on all-in-one image restoration: Taxonomy, evaluation and future trends},
  author={Jiang, Junjun and Zuo, Zengyuan and Wu, Gang and Jiang, Kui and Liu, Xianming},
  journal=TPAMI,
  year={2025}
}

@inproceedings{zamir2021multi,
  title={Multi-stage progressive image restoration},
  author={Zamir, Syed Waqas and Arora, Aditya and Khan, Salman and Hayat, Munawar and Khan, Fahad Shahbaz and Yang, Ming-Hsuan and Shao, Ling},
  booktitle=CVPR,
  pages={14821--14831},
  year={2021}
}

@inproceedings{conde2024instructir,
  title={InstructIR: High-Quality Image Restoration Following Human Instructions},
  author={Conde, Marcos V and Geigle, Gregor and Timofte, Radu},
  booktitle=ECCV,
  year={2024}
}

@inproceedings{nah2017deep,
  title={Deep multi-scale convolutional neural network for dynamic scene deblurring},
  author={Nah, Seungjun and Hyun Kim, Tae and Mu Lee, Kyoung},
  booktitle=CVPR,
  pages={3883--3891},
  year={2017}
}

@inproceedings{wei2018deep,
  title={Deep retinex decomposition for low-light enhancement},
  author={Wei, Chen and Wang, Wenjing and Yang, Wenhan and Liu, Jiaying},
  booktitle=BMVC,
  year={2018}
}

@inproceedings{chen2021hinet,
  title={Hinet: Half instance normalization network for image restoration},
  author={Chen, Liangyu and Lu, Xin and Zhang, Jie and Chu, Xiaojie and Chen, Chengpeng},
  booktitle=CVPR,
  pages={182--192},
  year={2021}
}

@inproceedings{mou2022deep,
  title={Deep generalized unfolding networks for image restoration},
  author={Mou, Chong and Wang, Qian and Zhang, Jian},
  booktitle=CVPR,
  pages={17399--17410},
  year={2022}
}

@inproceedings{zamir2020learning,
  title={Learning enriched features for real image restoration and enhancement},
  author={Zamir, Syed Waqas and Arora, Aditya and Khan, Salman and Hayat, Munawar and Khan, Fahad Shahbaz and Yang, Ming-Hsuan and Shao, Ling},
  booktitle=ECCV,
  pages={492--511},
  year={2020}
}

@inproceedings{liang2021swinir,
  title={Swinir: Image restoration using swin transformer},
  author={Liang, Jingyun and Cao, Jiezhang and Sun, Guolei and Zhang, Kai and Van Gool, Luc and Timofte, Radu},
  booktitle=ICCVW,
  pages={1833--1844},
  year={2021}
}

@article{cheng2024continual,
  title={Continual all-in-one adverse weather removal with knowledge replay on a unified network structure},
  author={Cheng, De and Ji, Yanling and Gong, Dong and Li, Yan and Wang, Nannan and Han, Junwei and Zhang, Dingwen},
  journal=TMM,
  year={2024}
}

@article{shin2021region,
  title={Region-based dehazing via dual-supervised triple-convolutional network},
  author={Shin, Joongchol and Park, Hasil and Paik, Joonki},
  journal=TMM,
  volume={24},
  pages={245--260},
  year={2021}
}

@inproceedings{chen2022simple,
  title={Simple baselines for image restoration},
  author={Chen, Liangyu and Chu, Xiaojie and Zhang, Xiangyu and Sun, Jian},
  booktitle=ECCV,
  pages={17--33},
  year={2022}
}

@article{fan2019general,
  title={A general decoupled learning framework for parameterized image operators},
  author={Fan, Qingnan and Chen, Dongdong and Yuan, Lu and Hua, Gang and Yu, Nenghai and Chen, Baoquan},
  journal=TPAMI,
  volume={43},
  number={1},
  pages={33--47},
  year={2019}
}

@inproceedings{valanarasu2022transweather,
  title={Transweather: Transformer-based restoration of images degraded by adverse weather conditions},
  author={Valanarasu, Jeya Maria Jose and Yasarla, Rajeev and Patel, Vishal M},
  booktitle=CVPR,
  pages={2353--2363},
  year={2022}
}

@inproceedings{liu2022tape,
  title={Tape: Task-agnostic prior embedding for image restoration},
  author={Liu, Lin and Xie, Lingxi and Zhang, Xiaopeng and Yuan, Shanxin and Chen, Xiangyu and Zhou, Wengang and Li, Houqiang and Tian, Qi},
  booktitle=ECCV,
  pages={447--464},
  year={2022}
}

@inproceedings{zhang2023ingredient,
  title={Ingredient-oriented multi-degradation learning for image restoration},
  author={Zhang, Jinghao and Huang, Jie and Yao, Mingde and Yang, Zizheng and Yu, Hu and Zhou, Man and Zhao, Feng},
  booktitle=CVPR,
  pages={5825--5835},
  year={2023}
}

@misc{franzen1999kodak,
  author= {Rich Franzen},
  title= {Kodak Lossless True Color Image Suite},
  howpublished={\url{http://r0k.us/graphics/kodak/}},
  year={1999},
  note={Online; accessed 24 October 2021}
}

@inproceedings{wang2022uformer,
  title={Uformer: A general u-shaped transformer for image restoration},
  author={Wang, Zhendong and Cun, Xiaodong and Bao, Jianmin and Zhou, Wengang and Liu, Jianzhuang and Li, Houqiang},
  booktitle=CVPR,
  pages={17683--17693},
  year={2022}
}

@inproceedings{yasarla2019uncertainty,
  title={Uncertainty guided multi-scale residual learning-using a cycle spinning cnn for single image de-raining},
  author={Yasarla, Rajeev and Patel, Vishal M},
  booktitle=CVPR,
  pages={8405--8414},
  year={2019}
}

@inproceedings{liang2022drt,
  title={Drt: A lightweight single image deraining recursive transformer},
  author={Liang, Yuanchu and Anwar, Saeed and Liu, Yang},
  booktitle=CVPR,
  pages={589--598},
  year={2022}
}

@inproceedings{zhang2021edge,
  title={Edge-oriented convolution block for real-time super resolution on mobile devices},
  author={Zhang, Xindong and Zeng, Hui and Zhang, Lei},
  booktitle=ACMMM,
  pages={4034--4043},
  year={2021}
}

@inproceedings{liu2019griddehazenet,
  title={Griddehazenet: Attention-based multi-scale network for image dehazing},
  author={Liu, Xiaohong and Ma, Yongrui and Shi, Zhihao and Chen, Jun},
  booktitle=ICCV,
  pages={7314--7323},
  year={2019}
}

@inproceedings{park2023all,
  title={All-in-one image restoration for unknown degradations using adaptive discriminative filters for specific degradations},
  author={Park, Dongwon and Lee, Byung Hyun and Chun, Se Young},
  booktitle=CVPR,
  pages={5815--5824},
  year={2023}
}

@inproceedings{
luo2024controlling,
title={Controlling Vision-Language Models for Multi-Task Image Restoration},
author={Ziwei Luo and Fredrik K. Gustafsson and Zheng Zhao and Jens Sj{\"o}lund and Thomas B. Sch{\"o}n},
booktitle=ICLR,
year={2024}
}

@inproceedings{ding2021repvgg,
  title={Repvgg: Making vgg-style convnets great again},
  author={Ding, Xiaohan and Zhang, Xiangyu and Ma, Ningning and Han, Jungong and Ding, Guiguang and Sun, Jian},
  booktitle=CVPR,
  pages={13733--13742},
  year={2021}
}

@inproceedings{ding2022scaling,
  title={Scaling up your kernels to 31x31: Revisiting large kernel design in cnns},
  author={Ding, Xiaohan and Zhang, Xiangyu and Han, Jungong and Ding, Guiguang},
  booktitle=CVPR,
  pages={11963--11975},
  year={2022}
}

@inproceedings{vasu2023mobileone,
  title={Mobileone: An improved one millisecond mobile backbone},
  author={Vasu, Pavan Kumar Anasosalu and Gabriel, James and Zhu, Jeff and Tuzel, Oncel and Ranjan, Anurag},
  booktitle=CVPR,
  pages={7907--7917},
  year={2023}
}

@inproceedings{zhang2025cvpr,
  title={Subspace Constraint and Contribution Estimation for Heterogeneous Federated Learning},
  author={Zhang, Xiangtao and Li, Sheng and Li, Ao and Liu, Yipeng and Zhang, Fan and Zhu, Ce and Zhang, Le},
  booktitle=CVPR,
  pages={1--1},
  year={2025}
}

@inproceedings{dong2023multi,
  title={Multi-scale residual low-pass filter network for image deblurring},
  author={Dong, Jiangxin and Pan, Jinshan and Yang, Zhongbao and Tang, Jinhui},
  booktitle={Proceedings of the IEEE/CVF International Conference on Computer Vision},
  pages={12345--12354},
  year={2023}
}

@inproceedings{yang2024all,
  title={All-in-one medical image restoration via task-adaptive routing},
  author={Yang, Zhiwen and Chen, Haowei and Qian, Ziniu and Yi, Yang and Zhang, Hui and Zhao, Dan and Wei, Bingzheng and Xu, Yan},
  booktitle={International Conference on Medical Image Computing and Computer-Assisted Intervention},
  pages={67--77},
  year={2024},
  organization={Springer}
}

@article{kong2024towards,
  title={Towards Effective Multiple-in-One Image Restoration: A Sequential and Prompt Learning Strategy},
  author={Kong, Xiangtao and Dong, Chao and Zhang, Lei},
  journal={arXiv preprint arXiv:2401.03379},
  year={2024}
}

@inproceedings{zhang2017learning,
  title={Learning deep {CNN} denoiser prior for image restoration},
  author={Zhang, Kai and Zuo, Wangmeng and Gu, Shuhang and Zhang, Lei},
  booktitle=CVPR,
  year={2017}
}

@inproceedings{cui2025adair,
      title={AdaIR: Adaptive All-in-One Image Restoration via Frequency Mining and Modulation}, 
      author={Cui, Yuning and Zamir, Syed Waqas and Khan, Salman and Knoll, Alois and Shah, Mubarak and Khan, Fahad Shahbaz},
      booktitle=ICLR,
      year={2025}
}

@inproceedings{guo2024mambair,
  title={Mambair: A simple baseline for image restoration with state-space model},
  author={Guo, Hang and Li, Jinmin and Dai, Tao and Ouyang, Zhihao and Ren, Xudong and Xia, Shu-Tao},
  booktitle=ECCV,
  pages={222--241},
  year={2024}
}

@article{li2023prompt,
  title={Prompt-in-prompt learning for universal image restoration},
  author={Li, Zilong and Lei, Yiming and Ma, Chenglong and Zhang, Junping and Shan, Hongming},
  journal={arXiv preprint arXiv:2312.05038},
  year={2023}
}

@article{wang2024gridformer,
  title={Gridformer: Residual dense transformer with grid structure for image restoration in adverse weather conditions},
  author={Wang, Tao and Zhang, Kaihao and Shao, Ziqian and Luo, Wenhan and Stenger, Bjorn and Lu, Tong and Kim, Tae-Kyun and Liu, Wei and Li, Hongdong},
  journal=IJCV,
  pages={1--23},
  year={2024}
}

@article{cui2023image,
  author={Cui, Yuning and Ren, Wenqi and Cao, Xiaochun and Knoll, Alois},
  journal=TPAMI, 
  title={Image Restoration via Frequency Selection}, 
  year={2024},
  volume={46},
  number={2},
  pages={1093-1108}
}

@article{liu2024pasta,
  title={PASTA: Towards Flexible and Efficient HDR Imaging Via Progressively Aggregated Spatio-Temporal Aligment},
  author={Liu, Xiaoning and Li, Ao and Wu, Zongwei and Du, Yapeng and Zhang, Le and Zhang, Yulun and Timofte, Radu and Zhu, Ce},
  journal={arXiv preprint arXiv:2403.10376},
  year={2024}
}

@article{DnCNN,
  author={Zhang, Kai and Zuo, Wangmeng and Chen, Yunjin and Meng, Deyu and Zhang, Lei},
  journal=TIP, 
  title={Beyond a Gaussian Denoiser: Residual Learning of Deep CNN for Image Denoising}, 
  year={2017},
  volume={26},
  number={7},
  pages={3142-3155}
}

@article{FFDNetPlus,
  author={Zhang, Kai and Zuo, Wangmeng and Zhang, Lei},
  journal=TIP, 
  title={FFDNet: Toward a Fast and Flexible Solution for CNN-Based Image Denoising}, 
  year={2018},
  volume={27},
  number={9},
  pages={4608-4622}
}

@article{zhang2025perceive,
  title={Perceive-ir: Learning to perceive degradation better for all-in-one image restoration},
  author={Zhang, Xu and Ma, Jiaqi and Wang, Guoli and Zhang, Qian and Zhang, Huan and Zhang, Lefei},
  journal=TIP,
  year={2025},
  publisher={IEEE}
}

@inproceedings{helou2020stochastic,
  title={Stochastic frequency masking to improve super-resolution and denoising networks},
  author={Helou, Majed El and Zhou, Ruofan and S{\"u}sstrunk, Sabine},
  booktitle={Proceedings of the European Conference on Computer Vision (ECCV)},
  pages={749--766},
  year={2020}
}

@inproceedings{qiu2019embedded,
  title={Embedded block residual network: A recursive restoration model for single-image super-resolution},
  author={Qiu, Yajun and Wang, Ruxin and Tao, Dapeng and Cheng, Jun},
  booktitle={Proceedings of the IEEE/CVF Conference on Computer Vision (ICCV)},
  pages={4180--4189},
  year={2019}
}

@inproceedings{magid2021dynamic,
  title={Dynamic high-pass filtering and multi-spectral attention for image super-resolution},
  author={Magid, Salma Abdel and others},
  booktitle={Proceedings of the IEEE/CVF Conference on Computer Vision (ICCV)},
  pages={4288--4297},
  year={2021}
}

@inproceedings{xie2021learning,
  title={Learning frequency-aware dynamic network for efficient super-resolution},
  author={Xie, Wen and Song, Dehua and Xu, Chunjing and Xu, Chuan and Zhang, Hui and Wang, Yi},
  booktitle={Proceedings of the IEEE/CVF Conference on Computer Vision (ICCV)},
  pages={4308--4317},
  year={2021}
}

@inproceedings{zhong2018joint,
  title={Joint sub-bands learning with clique structures for wavelet domain super-resolution},
  author={Zhong, Zhisheng and Shen, Tiancheng and Yang, Yibo and Lin, Zhouchen and Zhang, Chao},
  booktitle={Proceedings of the Conference and Workshop on Neural Information Processing Systems (NeurIPS)},
  pages={165--175},
  year={2018}
}

@article{li2025exploring,
  author={Li, Ao and Zhang, Le and Liu, Yun and Zhu, Ce},
  journal={IEEE Transactions on Pattern Analysis and Machine Intelligence}, 
  title={Exploring Frequency-Inspired Optimization in Transformer for Efficient Single Image Super-Resolution}, 
  year={2025},
  volume={47},
  number={4},
  pages={3141-3158},
}

@article{liu2021multi,
  title={Multi-scale grid network for image deblurring with high-frequency guidance},
  author={Liu, Yang and Fang, Faming and Wang, Tingting and Li, Juncheng and Sheng, Yun and Zhang, Guixu},
  journal={IEEE Transactions on Multimedia},
  volume={24},
  pages={2890--2901},
  year={2021}
}

@inproceedings{yu2022frequency,
  title={Frequency and spatial dual guidance for image dehazing},
  author={Yu, Hu and Zheng, Naishan and Zhou, Man and Huang, Jie and Xiao, Zeyu and Zhao, Feng},
  booktitle={Proceedings of the European Conference on Computer Vision (ECCV)},
  pages={181--198},
  year={2022}
}

@inproceedings{luo2023restoration,
  title={Restoration of Multiple Image Distortions using a Semi-dynamic Deep Neural Network},
  author={Luo, Hongming and Zhou, Fei and Zhou, Zehong and Lam, Kin-Man and Qiu, Guoping},
  booktitle={Proceedings of the ACM International Conference on Multimedia (ACMMM)},
  pages={7871--7880},
  year={2023}
}

@inproceedings{houlsby2019parameter,
  title={Parameter-efficient transfer learning for NLP},
  author={Houlsby, Neil and Giurgiu, Andrei and Jastrzebski, Stanislaw and Morrone, Bruna and De Laroussilhe, Quentin and Gesmundo, Andrea and Attariyan, Mona and Gelly, Sylvain},
  booktitle={Proceedings of the International Conference on Machine Learning (ICML)},
  pages={2790--2799},
  year={2019}
}

@inproceedings{chen2020dynamic,
  title={Dynamic convolution: Attention over convolution kernels},
  author={Chen, Yinpeng and Dai, Xiyang and Liu, Mengchen and Chen, Dongdong and Yuan, Lu and Liu, Zicheng},
  booktitle={Proceedings of the IEEE/CVF Conference on Computer Vision and Pattern Recognition (CVPR)},
  pages={11030--11039},
  year={2020}
}

@article{lihe2025ada4dir,
  title={Ada4DIR: An adaptive model-driven all-in-one image restoration network for remote sensing images},
  author={Lihe, Ziyang and Yuan, Qiangqiang and He, Jiang and Jin, Xianyu and Xiao, Yi and Chen, Yuzeng and Shen, Huanfeng and Zhang, Liangpei},
  journal={Information Fusion},
  volume={118},
  pages={102930},
  year={2025}
}

@article{chen2025profit,
  title={ProFiT: A prompt-guided frequency-aware filtering and template-enhanced interaction framework for hyperspectral video tracking},
  author={Chen, Yuzeng and Yuan, Qiangqiang and Tang, Yuqi and Wang, Xin and Xiao, Yi and He, Jiang and Lihe, Ziyang and Jin, Xianyu},
  journal={ISPRS Journal of Photogrammetry and Remote Sensing},
  volume={226},
  pages={164--186},
  year={2025}
}

@ARTICLE{zhang2025uniuir,
  author={Zhang, Xu and Zhang, Huan and Wang, Guoli and Zhang, Qian and Zhang, Lefei and Du, Bo},
  journal={IEEE Transactions on Image Processing}, 
  title={UniUIR: Considering Underwater Image Restoration as an All-in-One Learner}, 
  year={2025},
  volume={34},
  number={},
  pages={6963-6977}}
\end{document}